\documentclass[Times1COL]{WileyNJDv5} 

\RequirePackage{amsthm,amsmath,amsfonts,amssymb}
\RequirePackage{graphicx}

\usepackage{float}
\usepackage{comment}
\usepackage{mathtools}  
\usepackage{mathrsfs}

\usepackage{array}
\usepackage{booktabs}
\usepackage{multirow}
\usepackage{multirow}
\usepackage{subfig}
\usepackage{tabularx}
\usepackage[format=plain,font=normalsize,labelfont=bf]{caption}

\usepackage{stmaryrd}

\DeclareFontFamily{U}{zapfc}{\skewchar \font =45}
\DeclareFontShape{U}{zapfc}{m}{n}{<-> pzcmi}{}
\DeclareMathAlphabet{\mathcalzapf}{U}{zapfc}{m}{n}

\newcolumntype{L}{>{$}l<{$}} 
\newcolumntype{R}{>{$}r<{$}} 
\newcolumntype{C}{>{$}c<{$}} 

\DeclareMathOperator{\KL}{KL}
\DeclareMathOperator{\diag}{diag}

\DeclareMathOperator{\softmax}{softmax}
\DeclareMathOperator{\argmax}{argmax}

\DeclareMathOperator{\relu}{ReLU}

\articletype{Article Type}%

\startpage{1}

\begin{document}

\title{The Deep Latent Position Topic Model for Clustering and Representation of Networks with Textual Edges}

\author[1]{Rémi Boutin}

\author[2]{Pierre Latouche}

\author[3]{Charles Bouveyron}

\authormark{BOUTIN, LATOUCHE AND BOUVEYRON}
\titlemark{The Deep Latent Position Topic Model for Clustering and Representation of Networks with Textual Edges}

\address[1]{\orgdiv{Laboratoire MAP5, CNRS, UMR 8145}, \\ \orgname{Université Paris Cité}, \orgaddress{\state{Paris}, \country{France}}}

\address[2]{\orgdiv{Laboratoire LMBP, CNRS, UMR 6620}, \\ \orgname{Université Clermont Auvergne}, \orgaddress{\state{Aubière}, \country{France}}}

\address[3]{\orgdiv{Laboratoire J.A.Dieudonné, CNRS, Maasai team}, \orgname{Université Côte d'Azur, INRIA}, \orgaddress{\state{Nice}, \country{France}}}

\corres{Corresponding author Rémi Boutin, \email{remi.boutin.stat@gmail.com}}

\presentaddress{Université Paris Cité, Campus Saint-Germain, MAP5, 45 rue des Saints-Pères, 75006 Paris, France.}


\abstract[Abstract]{Numerical interactions leading to users sharing textual content published by others are naturally represented by a network where the individuals are associated with the nodes and the exchanged texts with the edges. To understand those heterogeneous and complex data structures, clustering nodes into homogeneous groups as well as rendering a comprehensible visualisation of the data is mandatory. To address both issues, we introduce Deep-LPTM, a model-based clustering strategy relying on a variational graph auto-encoder approach as well as a probabilistic model to characterise the topics of discussion. Deep-LPTM allows to build a joint representation of the nodes and of the edges in two embeddings spaces. The parameters are inferred using a variational inference algorithm. We also introduce IC2L, a model selection criterion specifically designed to choose models with relevant clustering and visualisation properties. An extensive benchmark study on synthetic data is provided. In particular, we find that Deep-LPTM better recovers the partitions of the nodes than the state-of-the art ETSBM and STBM. Eventually, the emails of the Enron company are analysed and visualisations of the results are presented, with meaningful highlights of the graph structure. }



\maketitle

\renewcommand\thefootnote{}

\renewcommand\thefootnote{\fnsymbol{footnote}}
\setcounter{footnote}{1}

	\section{Introduction and related work}

Numerical interactions between individuals often imply the creation of texts. For instance, on social media such as Twitter, it is possible to publish some content, a tweet or a post, that will in turn be republished, or re-twitted, by other accounts. Also, it is possible to mention another account directly in the publication. In the same way, the exchange of mails between collaborators can be seen as connections between accounts exchanging documents. Both examples can be represented by a network with the nodes corresponding to the accounts, and the edges to the exchanged texts. Such data structure is particularly difficult to apprehend, due to the heterogeneity and the volume of the data. One solution is to cluster homogeneous nodes into groups to obtain intelligible and useful information. However, very few methods performing node clustering are actually able to simultaneously exploit both the texts present on the edges, and the connections.

In the following, we first present the advancements in the statistical network analysis field. We then review some of the core probabilistic models that can capture the main topics in a corpus of texts. Eventually, we close this section with models considering both texts and networks to cluster nodes, before introducing the contribution of the present work.

\paragraph*{Statistical network analysis}
Network analysis first started with heuristic based methods. To obtain uncertainty estimates as well as more robust models, probabilistic modelling was developed in the second half of the 20th century. Statistical network analysis aims at capturing information from the network structure and making it intelligible for Human beings. One of the core model to perform model-based clustering on networks has been the stochastic block model (SBM) \cite{holland1983stochastic, snijders1997estimation, daudin2008mixture}. It assumes that the nodes are grouped into clusters and that the edges are independent of the nodes given their respective clusters. This model allows to discover any connectivity pattern. Another line of work has focused on obtaining relevant representations of the data in a Euclidean space. For instance, the latent position model (LPM, \cite{hoff2002latent}) obtains an informative representation of the network in a Euclidean space by assuming that each node can be modelled by a Gaussian variable in a low dimensional space. LPM was extended to the latent position clustering model (LPCM, \cite{handcock2007model}). The authors assumes that the latent node positions are sampled from a mixture of normal distributions. This extension  incorporates the clustering within the model and therefore combines the clustering task with the network representation. A review of standard models for statistical network analysis can be found in \cite{snijders2011statistical} and \cite{bouveyron2019model}.

\paragraph*{Deep probabilistic model for graph clustering}
Deep neural networks (DNN) have been introduced to perform clustering in the Euclidean setting, as described in \cite{aljalbout2018clustering}. The use of DNN, as a parametric function to encode data into a latent representation, has soared thanks to the variational auto encoders (VAE). This method rely on variational Bayes inference and the reparametrisation trick \cite{kingma2014autoencoding, rezende2014stochastic}. The interested reader can refer to \cite{zhang2018network}. Unfortunately, most traditional VAE can only be applied to Euclidean data, and, in particular, cannot be employed for network analysis. This issue was solved with the variational graph auto-encoder (VGAE, \cite{kipf2016variational}) that first combined the advancement in graph neural networks, namely the graph convolutional networks (GCN, \cite{kipf2016semi, henaff2015deep}) and deep probabilistic models. VGAE improves upon state-of-the-art methods for the link prediction task and is capable of using node features. In \cite{pan2018adversarially}, the authors regularise VGAE by using an adversarial inference strategy to improve the results. Unfortunately, those models rely on external methods, such as k-means, on the posterior node representations, to achieve node clustering. As an answer to this limit, \cite{mehta2019stochastic}  presented an extension of the overlapping stochastic block model \cite{latouche2011overlapping}. They proposed to encode the node embeddings with a NN. Lately, \cite{liang2022deep} combined the two and introduced the deep latent position model (Deep-LPM). Indeed, the model relies on VGAE and assumes that the latent representations are distributed according to a mixture of Gaussians, depending on the node cluster memberships. Their results advocate the incorporation of the clustering into the latent representations.

\paragraph*{Topic Modelling}
Like network analysis, the field of topic modelling was first developed using heuristics and was only  studied from a statistical perspective at the end of the 20th century. \cite{papadimitriou98latentsemantic} provided statistical foundations for the latent semantic index (LSI) and \cite{hofmann1999plsi} introduced a mixture model to represent the vocabulary distribution where each component modelled a specific topic. Ultimately, the latent Dirichlet allocation model (LDA) was proposed by \cite{blei2003latent}. They assumed that the topic proportions of the documents are sampled from a Dirichlet distribution. Many extensions of LDA have been proposed, including the use of a normal-logistic prior to better model the correlations between topics \cite{blei2006correlated}. In the end, the gap between deep generative modelling and topic models was bridged by \cite{srivastava2017autoencoding} who introduced a variational distribution parametrised by a DNN. More recently, the embedded topic model (ETM, \cite{dieng2020topic}). allowed to use embeddings in order to represent both the words and the topics in the same vector space. They are used as part of the decoder and can be pre-trained on large datasets to incorporate semantic meaning, as illustrated in the original paper with examples relying on continuous bag of words \cite{mikolov2013efficient}. 

\paragraph*{Joint analysis of network and texts}
While both topic modelling and statistical network analysis gave rise to many publications over the last 20 years, only few works have combined the two approaches.  The community-user topic model (CUT, \cite{zhou2006probabilistic}) added a latent variable to the author-topic model (AT, \cite{Rosenzvi2004author} to represent the communities either as a set of co-authors or as a set of topics. Thereafter, the community-author-recipient-topic model (CART, \cite{Pathak08socialtopic}) used communities both at the document generation level and at the author and recipient generation level which corresponds to the network generation.  However, the high number of parameters combined with the inference based on a Gibbs sampler does not allow to scale those models to large datasets. The topic-link LDA, presented in \cite{liu2009topic}, also offers a joint-analysis of texts and links in a unified framework by conditioning the generation of a link on both the topics within the documents and the community of authors. The inference relies on a variational EM algorithm to scale the approach to large datasets. However, this method only deal with undirected networks. Finally, the topic-user-community models (TUCM) was introduced in \cite{sachan2012using} and was able to discover topic-meaningful communities. The main feature of this model was its capacity to incorporate different types of interactions, well-suited for social networks applications. The inference relied on Gibbs sampling approach which can be limiting when dealing with large datasets. More recently, the stochastic topic block model (STBM) presented in \cite{bouveyron2018stochastic} was the first model to handle the simultaneous clustering of nodes and edges while keeping the inference tractable to large datasets thanks to a variational classification EM based inference. This model was extended in \cite{berge2019latent} for the simultaneous clustering of bipartite networks with textual edges. It was also adapted for dynamic networks in \cite{corneli2019dynamic}. Recently, \cite{boutin2022embedded} introduced the embedded topic in the stochastic block model (ETSBM)  to incorporate embeddings and semantic meanings into the word representations. However, despite their modelling qualities, all these approaches do not allow for the construction of a meaningful representation of the data which characterises both the connections and the documents exchanged. 

\paragraph*{Main contributions and organisation of the paper}
\begin{figure}
	\centering
	\includegraphics[width=1\linewidth]{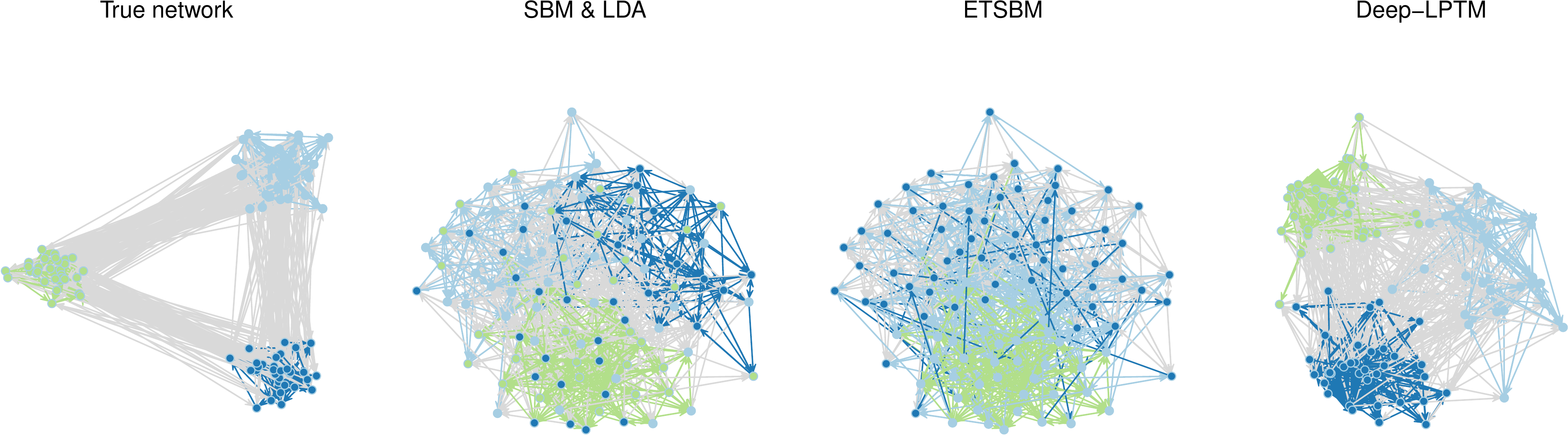}
	\caption{Illustration of Deep-LPTM main contributions on a synthetic network.
		The colours of the nodes and of the edges denote the node clusters and the main topics of the corresponding documents respectively. The true node partition as well as the true topics of the documents are represented on the left-hand side. The three other figures are based on the respective results of SBM and LDA (second figure), ETSBM (third figure) and Deep-LPTM (fourth figure). Only Deep-LPTM is able to provide node positions incorporating information about the network structure as well as the documents content. The four figures were obtained with the \textit{gplot} function from the \textit{sna} library \cite{krivitsky2022statnet}. While the first network is plotted manually to highlight the generative structure, the two networks in the middle are based on the Fruchterman-Reingold algorithm while the fourth graph uses the node positions estimated by Deep-LPTM.  Contrary to former methods STBM and ETSBM, which require the use of an exterior method, our methodology provides an end-to-end method tailored for representing networks with clustering properties. }
	\label{fig:introductionexamplewithoutcoordnew}
\end{figure}
The model proposed in the present paper is the first to simultaneously cluster the nodes of a network, uncover the topics in the texts exchanged between the nodes and to output a representation of both the topics and the edges in a Euclidean space. To this aim, \textbf{(i)} we propose a generative model assuming that each node and each edge is represented in a latent space by a mixture of Gaussians. By doing so, we incorporate the clustering in the generative process such that each mixture component models either a cluster of nodes or the documents exchanged between a pair of clusters of nodes. Moreover, our model distinguishes itself from former methods by allowing each node or edge to be represented by a latent position and not only by the group it belongs to. This is illustrated in Figure \ref{fig:introductionexamplewithoutcoordnew}, where the first graph depicts the true node clusters and main topics of the corresponding documents of a simulated dataset.The second graph gives, for this dataset, the node clusters provided by SBM and the topics estimated by LDA. SBM does not retrieve the true node partition and does not provide node positions to apprehend the results. An external algorithm, namely the Fruchterman-Reingold algorithm \cite{fruchterman1991graph}, considering the presence of connections only in the network, had to be used for graph representation. The third graph is based on state-of-the-art ETSBM and does not recover the node partition either. Moreover, ETSBM is not able to render a comprehensive representation of the full network. Again, as for SBM, the Fruchterman-Reingold algorithm had to be used for graph representation Finally, the figure on the right-hand side presents the Deep-LPTM results. As opposed to the previous methods, Deep-LPTM is able to gather the information about the network structure as well as the exchanged documents into the node positions while finding the true node partition and topics. A representation of the graph is directly obtained by the estimation procedure such that no external graph representation algorithm is needed. As we shall see, the node positions are computed by considering both the connections and the content of the corresponding documents. \textbf{(ii)} We derive a two-stage variational expectation-maximisation (VEM) algorithm for estimating the model parameters as well as the posterior parameters of the latent positions. The first stage relies on analytical formulas to update the cluster probabilities as well as the mixture parameters. The second stage uses a stochastic gradient descent algorithm to update the expected lower bound with respect to the VGAE parameters and the deep topic model parameters. In particular, the deep topic model can make the best out of pre-trained embeddings. Thus, introducing semantic meaning into the word representations is possible as well as learning the representation from scratch. \textbf{(iii)} In order to choose the relevant numbers of clusters and latent space dimensions, we introduce the integrated classification and latent likelihood (IC2L) for model selection. It extends the integrated classification likelihood criterion, which was conceived for mixture models, to account for the latent representations of the nodes and of the edges. The criterion relevance is strongly upheld by the evaluation on synthetic data as well as the provided real-word use case. Moreover, by selecting a low dimensions regarding the node embedding space,  IC2L praises for models with a strong and direct capacity of representation. 

We close this section with the organisation of the paper. In Section \ref{sec:model}, we present the assumptions concerning the generation of the data. The inference as well as the model selection criterion are presented in Section \ref{sec:inference}. In Section \ref{sec:numerical_xp}, Deep-LPTM and the impact of the initialisation are evaluated on synthetic data. An extensive benchmark study against state-of-the-art methods is also provided. Eventually, the emails of the Enron company are analysed with Deep-LPTM in Section \ref{sec:real_world_dataset}. The results as well as the visualisations are presented to illustrate the ease of interpretation of the model outcomes.

\subsection{Notations}
In this paper, we are interested in data represented by a graph $\mathcal{G} := \{ \mathcal{V}, \mathcal{E}\}$ where $\mathcal{V}=\{1, \dots, N\}$ denotes the set of vertices. The set $\mathcal{E}$ denotes the edges between the nodes with $M = | \mathcal{E}|$ the number of edges. We focus on binary a adjacency matrix $A \in \mathcal{M}_{N \times N}(\{0,1\})$ such that $A_{ij}$ equals $1$ if $(i,j) \in \mathcal{E}$, and $0$ otherwise. The graph is assumed to be directed and without any self loop. Therefore $A_{ii} = 0$ for all $i \in \mathcal{V}$. Finally, $Q$ denotes the number of clusters of nodes.

Each edge in the graph represents a textual document sent from one node to another. An edge from node $i$ to node $j$ exists or equivalently $(i,j) \in \mathcal{E}$ , if and only if node $i$ sent a textual document to node $j$, denoted $W_{ij}$. We use a bag-of-word representation of the texts where $W_{ij} = \left(W_{ij}^1, \dots,W_{ij}^V\right) \in \mathbb{N}^V$  denotes the vector of word occurrences in the document between nodes $i$ and $j$ such that $W_{ij}^v$ is the number of times word $v$ appears in the document, $M_{ij} = \sum_{v=1}^V W_{ij}^v$ is the total number of words in document $W_{ij}$ and $V$ the size of the vocabulary. Hence, $A_{ij}=1$ if $W_{ij}$ exists. The set of documents will be denoted $W := (W_{ij})_{(i,j) \in \mathcal{E}}$ and the number of topics is denoted by $K$. Eventually, the simplex of dimension $d$ will be denoted $\Delta_{d - 1}$.

\section{Model}\label{sec:model}

In the following, the assumptions about the graph generation as well as the hypothesis concerning the documents construction are presented.

\subsection{Graph generation}\label{sec:graph_generation}

Assuming that the number of clusters $Q$ is fixed before hand, each node $i$ is assumed to belong to a cluster, represented by the cluster membership variable $C_i$. The variables $C_i$, for any $i \in \mathcal{V}$, are assumed to be independent and identically distributed (i.i.d) according to a multinomial distribution such that for any node $ i \in \{1, \dots, N\}$:
\begin{align}\label{eq:distr_C}
	C_i \sim \mathcal{M}_{Q}(1, \pi),
\end{align}
with $\pi \in \Delta_{Q-1}$ and $C_i \in \{0,1\}^{Q}$ being one hot encoded so that $C_{iq} = 1$ if node $i$ belongs to cluster $q$ and $C_{iq}=0$ otherwise. Thus, denoting $C = (C_1, \dots, C_N)^{T} \in \mathcal{M}_{N \times Q}(\{0,1\})$ the cluster membership  matrix, we have:
\begin{align}\label{eq:proba_C}
	p(C \mid \pi) = \prod_{i=1}^N \prod_{q=1}^Q \pi_q^{C_{iq}}.
\end{align}
Moreover, given its cluster membership, the node $i$ is assumed to be represented by a Gaussian vector $Z_i$ in a $p$ dimensional latent space such that:
\begin{align}\label{eq:distr_Z}
	Z_i \mid C_{iq}=1 \sim \mathcal{N}\left( \mu_q, \sigma_q^2 I_p \right).
\end{align}
Eventually, the connection between two nodes is assumed to depend on the closeness of the node representations in the latent space. Therefore, denoting $\eta_{ij} := \kappa - \| Z_i - Z_j \|$,  which accounts for the opposite of the distance between the node latent representations, the probability for node $i$ to be connected to node $j$ is:
\begin{equation}\label{eq:proba_A}
	A_{ij} \mid Z_i, Z_j, \kappa \sim \mathcal{B}er\left(\frac{1}{1 + e^{- \eta_{ij}}} \right),
\end{equation}
where a logistic function is used as a link function. For the sake of brevity, we will denote $p_{ij} = \left(1 + e^{- \eta_{ij}} \right)^{-1}$. The probability of the entire adjacency matrix is given by:
\begin{equation*}
	\begin{split}
		p(A \mid Z, \kappa) & = \prod_{ i \neq j } p_{ij}^{A_{ij}} (1 - p_{ij})^{1 - A_{ij}}.
	\end{split}
\end{equation*}
Finally, the joint-distribution of the adjacency matrix, the latent node vectors, as well as the cluster memberships can be factorised as follow:
\begin{equation}
	p(A, Z, C \mid \kappa, \mu, \sigma, \pi) = p(A \mid Z, \kappa) p(Z \mid C, \mu, \sigma) p(C \mid \pi).
\end{equation}
where $\mu=(\mu_q)_q$ and $\sigma=(\sigma_q)_q$.
It is worth noticing that the model described in Equations \eqref{eq:distr_C}, \eqref{eq:distr_Z} and \eqref{eq:proba_A} corresponds to the latent position cluster model \cite{handcock2007model}. The fundamental difference with our approach for this part of the model will arise in the inference, as discussed in Section \ref{sec:inference}. 

\subsection{Generation of the texts on the edges}\label{sec:texts_generation}

At the core of our approach is the motivation to be able to use textual data  to obtain more homogeneous and meaningful clusters. 

To begin with, we make the assumption that each edge can be represented in a latent space by a Gaussian vector, depending only on the node cluster memberships. Thus, given $(C_i)_{i \in \mathcal{V}}$, the latent variables $Y_{ij}$ are assumed to be i.i.d such that: 
\begin{align}
	Y_{ij} \mid A_{ij} C_{iq} C_{jr} = 1 \sim \mathcal{N}( m_{qr} , s_{qr}^2 I_K), \ \forall (i,j) \in \mathcal{E},
\end{align}
where $m_{qr} \in \mathbb{R}^{K}, s_{qr} \in \mathbb{R^{+}}$. 

Moreover, we assume that the topic proportions of the document $W_{ij}$, denoted $\theta_{ij}$, can be deduced from the latent variables such that: 
\begin{align}
	\tilde{Y}_{ij} = \softmax(Y_{ij}),
\end{align}
where $\softmax(x) = \left( \sum_{f=1}^F e^{x_f} \right)^{-1} \left(e^{x_1}, \dots, e^{x_F} \right)^{\top}$ if $x \in \mathbb{R}^F$.

Hence, assuming that the documents are i.i.d  given their corresponding topic proportions, we have for any edge $(i,j) \in \mathcal{E}$:
\begin{align}
	W_{ij} \mid A_{ij}=1, \theta_{ij} \sim \mathcal{M}_{V} \left(M_{ij}, \beta^{\top} \theta_{ij} \right),
\end{align}
\begin{sloppypar}\noindent
	where  $\beta_k = \softmax(\rho^{\top} \alpha_k) \in \mathbb{R}^{V}$, $\beta = (\beta_1 \dots \beta_K)^{\top} \in \mathcal{M}_{K \times V}(\mathbb{R})$, $\rho \in \mathcal{M}_{L \times V}(\mathbb{R})$, $\alpha_k~\in~\mathbb{R}^L$ and $\alpha~=~\left( \alpha_1 \dots \alpha_K \right)~\in~\mathcal{M}_{L \times K}(\mathbb{R})$.
	Thus, denoting $m=(m_{qr})_{qr}$, $s=(s_{qr})_{qr}$, the joint likelihood of the texts, the latent representation of the documents, as well as the clusters memberships can be computed as:
\end{sloppypar}\noindent
\begin{align}
	p(W, Y \mid A, C, \rho, \alpha, m, s) = p(W \mid A, Y, \rho, \alpha) p(Y \mid  A, C, m, s),
\end{align}
where $m=(m_{qr})_{1 \leq q,r \leq Q}$ and $s=(s_{qr})_{1 \leq q,r \ Q}$. 

\begin{figure}
	\centering
	\includegraphics[width=0.8 \linewidth]{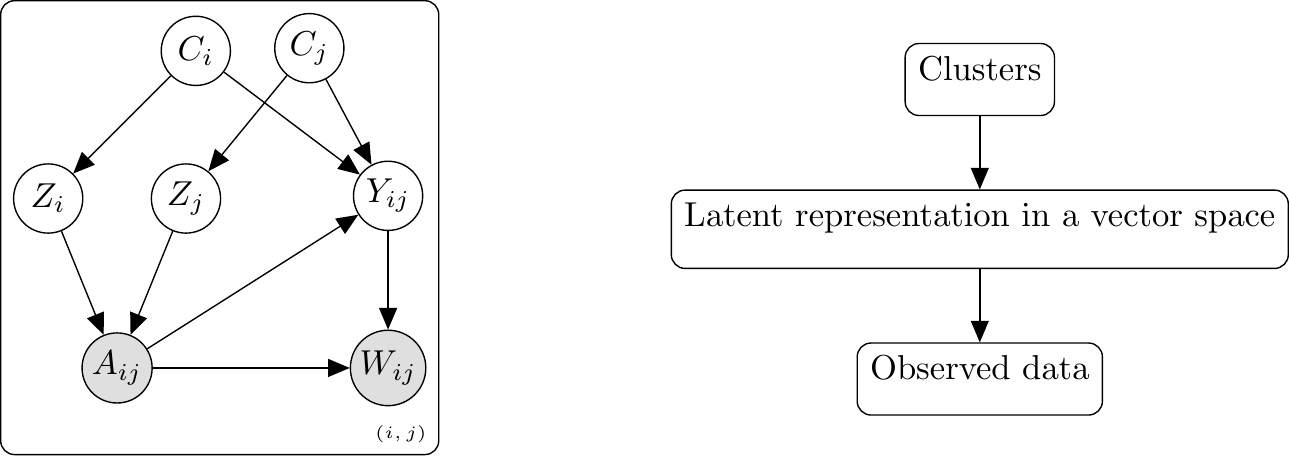}
	\caption{Graphical representation of the model without the parameters for the sake of clarity.}
	\label{graphical_model}
\end{figure}

Figure \ref{graphical_model} gives a graphical representation of the generative assumptions presented above. We omitted the parameters for the sake of clarity but the full version can be found in Appendix \ref{fig:graphical_model_full}. Interestingly enough, those assumptions can be linked to existing models as we shall see in the next section.

\subsection{Link with other models}
On the one hand, if the topic modelling alone is considered, restricting all topic proportions to be equal for all $(i,j)$ such that $C_{iq} C_{jr}~=~1$ corresponds to the text modelling in ETSBM \cite{boutin2022embedded}. In that sense, Deep-LPTM increases the freedom of each edge representation compared to ETSBM. Additionally, Deep-LPM corresponds to Deep-LPTM when the textual data present on the edges are disposed of (or LPCM if no GCN-based encoder is used in the inference strategy). Accordingly, Deep-LPTM prolongs Deep-LPM and LPCM to networks with textual data. On the other hand, discarding the information provided by the graph and the clustering of the nodes would correspond to ETM applied on the observed documents. Therefore, Deep-LPTM extends ETM to texts with a connectivity structure to improve the topic modelling. 

\section{Inference}\label{sec:inference}
In the next section, the inference of the model is presented as well as the model selection criterion.
\subsection{Likelihood}
\begin{sloppypar}\noindent
	In this work, we consider the marginal likelihood of the network and the texts for parameter estimation. The latent variables are denoted by $C=(C_i)_{i=1}^N$, $Z=(Z_i)_{i=1}^N$ and $Y = (Y_{ij})_{ (i, j) \in \mathcal{E} }$, and the set of parameters is $\Theta = \left\{ \pi, \mu, \sigma, \kappa, m, s, \alpha, \rho  \right\}$ .Thus, the marginal log-likelihood is given by:
\end{sloppypar}\noindent
\begin{align}\label{eq:log_likelihood}
	\mathcal{L}(\Theta; A,W) = \log  p(A, W \mid \Theta)  = \log \left( \sum_{C} \int_{Z} \int_{Y} p(A, W, C, Z, Y \mid \Theta) d Z d Y \right).
\end{align}
Unfortunately, this quantity is not tractable since the sum over $C$  requires to compute $Q^N$ terms. Besides, it involves integrals that cannot be computed analytically. Therefore, we choose to rely on a variational inference approach for approximation purposes.

\paragraph{Decomposition of the marginal log-likelihood}
\begin{sloppypar}
	For any distribution $R(C,Z,Y)$, the following decomposition holds:
\end{sloppypar}

\begin{align}\label{eq:elbo_split}
	\mathcal{L}(\Theta; A,W) = \mathscr{L}(R(\cdot); \Theta) + \KL( R(\cdot) || p(C,Z,Y \mid A, W)),
\end{align}
where
\begin{align}\label{eq:ELBO_not_decomposed}
	\mathscr{L}(R(\cdot); \Theta) = \mathbb{E}_{R}\left[ \log \frac{p(A,W, C, Z, Y\mid \Theta)}{R(C,Z,Y)}\right].
\end{align}
Since the Kullback-Leibler divergence is always positive in Equation \eqref{eq:elbo_split}, the ELBO $\mathscr{L}(R(\cdot); \Theta)$ is a lower bound of the marginal log-likelihood. Moreover, the closer $R(\cdot)$ is to the posterior distribution of the latent variables, in terms of Kullback-Leibler divergence, the closer the ELBO is to the marginal log-likelihood. Since the marginal log-likelihood does not depend on $R(\cdot)$, maximizing the ELBO with respect to $R(\cdot)$ is equivalent to minimizing the Kullback-Leibler divergence between $R(\cdot)$ and the posterior distribution. To make the ELBO tractable, we restrict the family of variational distributions by considering a mean field assumption. Moreover, the assumptions in Equations \eqref{eq:var_distr_clusters} to \eqref{eq:var_distr_Y} allow us to take full advantage of the SGD efficiency to optimise the variational distribution while keeping a high flexibility thanks to the neural network parametrisations of the distributions (see, section 2.5 and 2.6 of \cite{kingma2019introduction}). The assumptions are given by:
\begin{align}
	&R(C,Z,Y \mid A, W) = R(C)R(Z \mid A)R(Y \mid A, W), \label{eq:mean_field}\\
	&R(C)=\prod_{i=1}^N R_{\tau_i}(C_i) =\prod_{i=1}^N \mathcal{M}_{Q}(C_i; 1,\tau_i),  \label{eq:var_distr_clusters}\\
	&R(Z \mid A) = \prod_{i=1}^N R_{\phi_Z}(Z_i \mid A) = \prod_{i=1}^N  \mathcal{N}(Z_i; \mu_{\phi_Z}(A)_i, \sigma^2_{\phi_Z}(A)_i I_P),  \label{eq:var_distr_Z} \\
	& R(Y \mid A, W) = \prod_{i \neq j}  R_{\phi_Y}(Y_{ij} \mid W_{ij})^{A_{ij}} = \prod_{i \neq j}  \mathcal{N} \bigl(Y_{ij}; \mu_{\phi_Y}(W_{ij}), \diag\left(\sigma^2_{\phi_Y}(W_{ij}) \right) \bigr)^{A_{ij}}, \label{eq:var_distr_Y}
\end{align}
\begin{sloppypar}\noindent
	where $\tau = (\tau_i)_{i=1}^N$ with $\forall i \in \{1, \dots, N\}$, $\tau_i \in \Delta_{Q - 1}$.
	Moreover, in Equation \ref{eq:var_distr_Z}, the mapping $\mu_{\phi_Z}~\colon~\mathcal{M}_{N \times N}(\mathbb{R})~\mapsto~\mathcal{M}_{N \times P}(\mathbb{R})$ ($\sigma^2_{\phi_Z}~\colon~\mathcal{M}_{N \times N}(\mathbb{R})~\mapsto~(\mathbb{R}^{+})^N $ respectively) is the mapping normalising the adjacency matrix by its degree, $D^{-1/2} A D^{-1/2}$, and encoding the normalised adjacency matrix into the approximated posterior means (standard deviations) of the node latent positions. The diagonal matrix $D$ is filled with $D_{ii}$, the degree of node $i$, for all nodes. The two mappings $\mu_{\phi_Z}$ and $\sigma^2_{\phi_Z}$ rely on a GCN parametrised by $\phi_Z$  \cite{kipf2016variational}.
	Similarly, in Equation \ref{eq:var_distr_Y}, $\mu_{\phi_Y}~\colon~\mathcal{M}_{M \times V}(\mathbb{R})~\mapsto~\mathcal{M}_{M \times K}(\mathbb{R})$ ($\sigma^2_{\phi_Y}~\colon~\mathcal{M}_{M \times V}(\mathbb{R})~\mapsto~\mathcal{M}_{M \times K}(\mathbb{R}^+) $ respectively) encodes the documents into the approximated posterior means (standard deviations) of their corresponding latent vectors. However, the two functions rely on the ETM encoder with parameter $\phi_Y$ \cite{dieng2020topic}. In practice, in all the experiments we carried out, we used a feed-forward neural network to encode the documents, with three layers and 800 units on each layer. The first two layers are shared to encode the variances and the means while the last layer is specific to each vector. Regarding the encoder of the adjacency matrix, we rely on \cite{kipf2016variational} such that,  $\mu_{\phi_Z}(A) = \tilde{A} \relu(\tilde{A} W_0) W_{\mu}$ and $ \log \sigma^{2}_{\phi_Z}(A) = \tilde{A} \relu(\tilde{A}W_0) W_{\sigma}$, where $\tilde{A} = D^{-1/2} A D^{-1/2}$ and $\relu(x) = ( \max(0, x_1), \dots, \max(0, x_F))$ if $x \in \mathbb{R}^F$. $\mu_{\phi_Z}(\cdot)$ and $\log \sigma^{2}_{\phi_Z}(\cdot)$ share the first-layer parameter $W_{0} \in \mathcal{M}_{N \times D}$ with $D=10$ in all the experiments we carried out, and $W_{\mu}, W_{\sigma} \in \mathcal{M}_{D \times P}$. For the sake of brevity, we will take the exponential of the encoder of the log variance and consider $\sigma^{2}_{\phi_Z}(\cdot)$.
\end{sloppypar}

Thus, the ELBO can be decomposed as follow :
\begin{align}\label{eq:elbo_detailed}
	\mathscr{L}(R(\cdot); \Theta) &= \mathbb{E}_{R}\left[ \log p(A \mid Z, \kappa) \right] +\mathbb{E}_{R}\left[ \log p(W \mid A, Y, \rho, \alpha) \right] +\mathbb{E}_{R}\left[ \log p(C \mid \pi) \right] \nonumber\\
	&+\mathbb{E}_{R}\left[ \log p(Z \mid C, \mu, \sigma)\right]
	+\mathbb{E}_{R}\left[ \log p(Y \mid A, C, m, s) \right]  - \mathbb{E}_{R}\left[\log R(C) \right] \nonumber\\
	& -\mathbb{E}_{R}\left[ \log R(Z \mid A) \right] -\mathbb{E}_{R}\left[ \log R(Y \mid A, W)\right].
\end{align}
The computation of each term in Equation \eqref{eq:elbo_detailed} is detailed in Appendix \ref{sec:elbo_computation_details}. To optimise the ELBO, we propose here to alternate between closed form updates and stochastic gradient descent steps thanks to the results presented in the next section.

\subsection{Optimisation}

\subsubsection{Analytical updates} Given a variational distribution $R(\cdot)$ complying with Equations \eqref{eq:mean_field} to \eqref{eq:var_distr_Y}, the model parameters can be updated using Propositions \eqref{prop:nodes_param_update} and \eqref{prop:edges_param_update}.

\begin{proposition}\label{prop:nodes_param_update}
	Let $R(\cdot)$ be a variational distribution complying with Equations \eqref{eq:mean_field} to \eqref{eq:var_distr_Y}. The parameters of the nodes embeddings distributions maximising the ELBO are given by:
	\begin{align}
		\mu_q & = \frac{1}{N_q} \sum_{i=1}^N \tau_{iq}  \mu_{\phi_Z}(A)_{i},\\
		\sigma_q^2 & =  \frac{1}{p N_q} \sum_{i=1}^N \tau_{iq} \left( p \sigma_{\phi_Z}^2(A)_{i} + \| \mu_{\phi_Z}(A)_{i} - \mu_q \|_2^2 \right),
	\end{align}
	where $N_q = \sum_{i=1}^N \tau_{iq} $ is the posterior mean of the number of nodes in cluster $q$.
\end{proposition}
The proof is given in Appendix \ref{proof:Z_params_update}. Interestingly, this proposition states that the $\mu_q$ are the weighted mean of the (approximated) posterior mean nodes positions $\mu_{\phi_Z}$ provided by the DNN. It also indicates that the $\sigma_q$ are updated as the sum of two terms: the first one corresponds to a weighted mean of the posterior variances while the second one is the intra cluster variance weighted by the posterior clusters membership probabilities $\tau_i$.

\begin{proposition}\label{prop:edges_param_update}
	For a given variational distribution $R(\cdot)$ complying with Equations \eqref{eq:mean_field} to \eqref{eq:var_distr_Y}, with parameters $\tau,  \mu_{\phi_Y}, \sigma_{\phi_Y} $, the parameters of the edge embeddings distributions maximising the ELBO are given by:
	\begin{align}
		m_{qr} & = \frac{1}{N_{qr}} \sum_{i,j=1}^N A_{ij}\tau_{iq}\tau_{jr}  \mu_{\phi_Y}(W_{ij}),\\
		s_{qr}^2 & = \frac{1}{K N_{qr}} \sum_{i,j=1}^N A_{ij}\tau_{iq}\tau_{jr} \left[\sum_{k=1}^K \sigma_{\phi_Y}^2(W_{ij})_{k}  + \| \mu_{\phi_Y}(W_{ij}) - m_{qr} \|_2^2 \right],
	\end{align}
	where $N_{qr} = \sum_{i,j=1}^N A_{ij}\tau_{iq}\tau_{jr}$ denotes the expected number of documents sent from cluster $q$ to cluster $r$ under the approximated posterior distribution.
\end{proposition}
The proof is given in Appendix \ref{proof:Y_params_update}
The interpretation following Proposition \ref{prop:nodes_param_update} can  also be applied to Proposition \ref{prop:edges_param_update} while a second DNN is used. The main difference lies in the weighting that here corresponds to the probability that the pair of nodes composing each edge belong to a pair of cluster. For instance, for the edge $(i,j)$, the posterior probability that node $i$ belongs to cluster $q$ and node $j$ to cluster $r$ is given by  $\tau_{iq} \tau_{jr}$.

\subsubsection{The stochastic gradient descent}
The other model parameters  $\kappa, \rho$ and $\alpha$, and variational parameters $\phi_Z$ and $\phi_Y$ cannot be updated with analytical formulas because of the integral involving the variational distribution $R(\cdot)$ in the ELBO. In the following section, we aim at deriving estimates of the gradients of the ELBO with respect to these parameters, to perform stochastic gradient descent.

\paragraph{Model parameters}
The partial derivatives of the ELBO with respect to each parameter $\kappa, \rho$ and $\alpha$ are obtained thanks to Monte-Carlo estimates. For instance, the gradient of the ELBO with respect to $\kappa$ is estimated by:
\begin{align*}
	\frac{\partial }{\partial \kappa} \mathscr{L}(R(\cdot); \Theta) & = \frac{\partial }{\partial \kappa}  \mathbb{E}_{R} \left[  \log p(A \mid Z, \kappa)  \right] \\
	& =  \mathbb{E}_{R} \left[ \frac{\partial }{\partial \kappa} \log p(A \mid Z, \kappa)  \right] \\ 
	& \approx  \frac{1}{S} \sum_{s=1}^{S} \frac{\partial }{\partial \kappa} \log p(A \mid Z^{(s)}, \kappa),
\end{align*} 
where $Z^{(s)} = (Z_{1}^{(s)}, \dots,Z_{N}^{(s)} ) $ and $Z_{i}^{(s)} \overset{i.i.d}{\sim} \mathcal{N}(\mu_{\phi_Z}(A)_i, \sigma^2_{\phi_Z}(A)_i I_p)$. The same computations result in estimates for the partial derivatives of the ELBO with respect to $\rho$ and $\alpha$. In practice, we rely on the common practice in the field of VAE and set $S=1$. 

\paragraph{Variational parameters}
The last parameters to update are the variational parameters $\phi_{Y}$ and $\phi_{Z}$. Ideally, we would like to use the same computation as in the previous section. For instance, we would like to compute the following partial derivates of the ELBO with respect to $\phi_Z$:

\begin{align}\label{eq:before_reparam}
	\frac{\partial}{\partial \phi_{Z}} \mathscr{L}( R(\cdot); \Theta) & = \frac{\partial}{\partial \phi_{Z}} \mathbb{E}_{R}\left[   \log p(A \mid Z, \kappa) +  \log p(Z \mid C, \mu, \sigma) - \log R(Z) \right] \nonumber\\
	& = \frac{\partial}{\partial \phi_{Z}} \mathbb{E}_{R}\left[   \log p(A \mid Z, \kappa)] \right] \\
	& \phantom{ = } -  \sum_{i=1}^N \sum_{q=1}^Q \tau_{iq}  \frac{\partial}{\partial \phi_{Z}} \KL_{iq}^Z\left(\mu_{\phi_Z}(A)_{i}, \sigma_{\phi_Z}(A)_{i}, \mu_q, \sigma_{q}\right),
\end{align}
with 
\begin{align*}
	\KL_{iq}^Z\left(\mu_{\phi_Z}(A)_{i}, \sigma_{\phi_Z}(A)_{i}, \mu_q, \sigma_{q}\right) = & \log \frac{\sigma_q^{p}}{\sigma_{\phi_Z}(A)_{i}^{p}} - \frac{p}{2} \\
	& + \frac{ p \sigma_{\phi_Z}^2(A)_{i} + \| \mu_{\phi_Z}(A)_{i} - \mu_q \|_2^2 }{ 2 \sigma_q^{2}}.
\end{align*}
See Appendix \ref{eq:KL_nodes} for the computational details. Unfortunately, the expectation of the term on the left-hand side of Equation \eqref{eq:before_reparam}, is taken with respect to the variational distribution which depends on $\phi_Z$. Therefore, the derivation of this quantity is not straightforward. Thanks to \cite{kingma2014autoencoding} and \cite{rezende2014stochastic}, this difficulty can be tackled by using the reparametrisation trick. In particular, let $\varepsilon_i$ be a centred, normalised and $P$-dimensional Gaussian vector. Hence, the vectors $Z_i$ and $\displaystyle \mu_{\phi_{Z}}(A)_i \oplus \left[ \sigma_{\phi_Z}(A)_i \odot \varepsilon_i \right]$ have the same distribution. Therefore, the expectation can be taken with respect to $\epsilon = (\varepsilon_i)_{i=1\dots,N}$ which gives:
\begin{align*}
	\frac{\partial}{\partial \phi_{Z}} \mathbb{E}_{R}\left[   \log p(A \mid Z, \kappa)] \right]  =  \frac{\partial}{\partial \phi_{Z}} \mathbb{E}_{\varepsilon}\Bigl[ &  \log p\left(A \mid \mu_{\phi_{Z}}(A) \oplus \left[\sigma_{\phi_Z}(A) \odot \epsilon \right], \kappa \right) \Bigr]
\end{align*}
where $\displaystyle \mu_{\phi_{Z}}(A) \oplus \left[\sigma_{\phi_Z}(A) \odot \epsilon\right] = \Bigl(\mu_{\phi_{Z}}(A)_i \oplus \left[ \sigma_{\phi_Z}(A)_i \odot \varepsilon_i \right] \Bigr)_{i=1,\dots,N}$, $\odot$ denotes the Hadamard product and $\oplus$ the element-wise sum. A Monte-Carlo estimate of this quantity is derived by sampling $S$ centred and reduced $P$-dimensional Gaussian vectors $\epsilon_i^{(s)}$ with $s=1,\dots,S$, $i = 1 , \dots, N$ and with $\epsilon^{(s)} = (\varepsilon_i^{(s)})_{i=1,\dots,N}$. Plugging it back into \eqref{eq:before_reparam} gives the following estimate:
\begin{align*}
	\frac{\partial}{\partial \phi_{Z}} \mathscr{L}( R(\cdot); \Theta) \approx & \frac{1}{S} \sum_{s=1}^S \Bigl[  \frac{\partial}{\partial \phi_{Z}} \log p \left(A \mid \mu_{\phi_{Z}}(A) \oplus \left[ \sigma_{\phi_Z}(A) \odot \epsilon^{(s)} \right], \kappa \right) \Bigr] \\
	& -  \sum_{i=1}^N \sum_{q=1}^Q \tau_{iq}  \frac{\partial}{\partial \phi_{Z}} \KL_{iq}^Z\left(\mu_{\phi_Z}(A)_{i}, \sigma_{\phi_Z}^2(A)_{i}, \mu_q, \sigma_{q}^2\right) .
\end{align*}
The same derivation steps lead to a similar estimate for the partial derivatives of the ELBO with respect to  $\phi_{Y}$. Thanks to the low variances of the gradients estimated with the reparametrisation trick and to avoid increasing the computations, we use a sample size  $S=1$, as advised in the VAE literature. In addition, the computation of the partial derivatives is implemented with Pytorch automatic differentiation framework \cite{pytorch2019nips} to take fully advantage of the computational efficiency of GPUs. Moreover, we rely on the Adam optimiser \cite{kingma2014adam} to carry out the stochastic gradient descent with a learning rate of $0.002$ ($0.005$ respectively) for the optimiser of $\kappa$ and $\phi_Z$ ($\phi_Y$ respectively).

\subsection{Model selection}
To complete the inference, we present IC2L, a new model selection criterion accounting for both the clustering partition as well as the latent representations. In all previous sections, we considered the number of clusters $Q$, the number of topics $K$ and the dimension of the node latent space $P$ fixed before hand. In this section, we aim at selecting the triplet $(K,P,Q)$ that captures the most information out of the data without over-parametrisation. The integrated complete (or classification) likelihood \cite{biernacki2000assessing}, denoted ICL,  was introduced for mixture models and is now a common model selection criterion in this context.

First, considering a mixture model $\mathcalzapf{M}$, with observed data $X$, latent cluster memberships $C$, $Q$ clusters and distribution parameters in the set $\Theta$, the complete likelihood refers to $ p(X,C \mid \mathcalzapf{M}, Q, \Theta)$. This quantity depends on both the clustering and the model parameters. Then, to account for the uncertainty over the set of parameters and to penalise the model complexity, \cite{biernacki2000assessing} proposed to integrate over $\Theta$ and to evaluate the quantity $\log~p(X,C~\mid~\mathcalzapf{M}, Q) = \int_{\theta \in \Theta} \log p(X, C \mid  \mathcalzapf{M}, Q, \theta) d\theta$. Since the integral is not tractable for many statistical models, the authors relied on a BIC-like approximation of this quantity. In the present paper, we propose to extend ICL to include the evaluation of the node embeddings $Z$, as well as the edge embeddings $Y$. Denoting $ \mathcalzapf{M}$ the model presented in Sections \ref{sec:graph_generation} and \ref{sec:texts_generation}, we are interested in:
\begin{align}
	\log p(A,W,Z,Y,C \mid \mathcalzapf{M}, Q, K, P ) &= \log \int_{\theta} p(A, W, Z, Y, C \mid \theta, \mathcalzapf{M}, Q, K, P) p(\theta) d \theta. \label{eq:int_ic2l}
\end{align} 
Since this quantity is not tractable, we derive an estimate in the following proposition.
\begin{proposition}\label{prop:IC2L}
	\begin{sloppypar}\noindent
		Let us consider a model $\mathcalzapf{M}$, as described in Section \ref{sec:model}, with $Q$ denoting the number of clusters, $K$ the number of topics and $P$ the dimension of the node latent space.  In addition, let us assume that the prior of the model parameters fully factorises, as $p(\kappa, \pi, \mu, \sigma, m, s, \rho, \alpha) =  p(\kappa)  p(\pi) p(\mu) p( \sigma) p(m) p(s) p(\rho) p(\alpha)$. Then, the $	IC2L( \mathcalzapf{M}, Q, K, P, \hat{Z}, \hat{Y}, \hat{C})$ criterion, denoted $IC2L$, is given by:
	\end{sloppypar}\noindent
	\begin{align}
		IC2L & = \max_{\theta} \log p(A,W,\hat{Z},\hat{Y}, \hat{C} \mid \theta, \mathcalzapf{M}, Q, K, P )  - \Omega(\mathcalzapf{M}, Q, K, P) \nonumber\\
		& =  \max_{\kappa} \log p(A \mid \hat{Z}, \kappa, \mathcalzapf{M}) - \frac{ 1 }{2} \log(N (N- 1)) \nonumber\\
		& + \max_{\mu, \sigma} \log p (\hat{Z} \mid \hat{C}, \mu, \sigma, \mathcalzapf{M}, Q, P) - \frac{Q P + Q}{2} \log(N)\nonumber \\
		& + \max_{\rho, \alpha }\log p(W \mid A, \hat{Y}, \rho, \alpha,  \mathcalzapf{M}) - \frac{V L + K  L}{2} \log(M) \nonumber\\
		& + \max_{m, s} \log p(\hat{Y} \mid A, \hat{C}, m, s, \mathcalzapf{M}, K ) - \frac{ Q^2  K + Q^2  }{2} \log(M) \nonumber \\ 
		& + \max_{\pi }\log p(\hat{C} \mid \pi, \mathcalzapf{M}, Q) - \frac{ Q-1}{2} \log( N ), \label{eq:model_selection_dvt}
	\end{align}
	with $\hat{Z}, \hat{Y}$ and, $\hat{C}$ the maximum-a-posteriori estimates, and 
	\begin{equation}\label{eq:omega}
		\begin{split}
			\Omega(\mathcalzapf{M}, Q, K, P) = & \frac{1}{2}\log(N(N~-~1)) \\
			& +\frac{Q (P + 2) - 1}{2} \log(N) \\
			& +  \frac{L (V + K)+ Q^2  (K + 1) }{2} \log(M),
		\end{split}
	\end{equation}
	where each term in Equation \eqref{eq:omega} corresponds to the penalisation of the BIC-like approximation of the probabilities, as detailed in Equation \eqref{eq:model_selection_dvt}.
\end{proposition}
\begin{proof}
	See Appendix \ref{proof:IC2L}.
\end{proof}
Ultimately, the relevance of this criterion as well as the parameters estimation are assessed in the next section on synthetic data. Moreover, an extensive comparison with baseline methods is provided. 

\section{Numerical experiments}\label{sec:numerical_xp}
\begin{sloppypar}
	This section is dedicated to the assessment of the proposed methodology. We start with an introductory example to illustrate the results obtained with Deep-LPTM. We continue with an evaluation of the initialisation impact on our method. Then, we move on to Section \ref{sec:model_selection_xp} to provide numerical evidence of the robustness of IC2L against the dimensions of the parameter spaces. We close this section with a benchmark study to compare Deep-LPTM with the state-of-the-art ETSBM and STBM.\footnote{Our code is available at \href{https://plmlab.math.cnrs.fr/rboutin/deeplptm_package}{https://plmlab.math.cnrs.fr/rboutin/deeplptm\_package}.}
\end{sloppypar}

\subsection{Simulation settings}
To begin with, we introduce three simulation scenarios to be used for evaluating the methodology on different conditions detailed hereinafter.
\paragraph{Scenarios}

\begin{figure}
	\centering
	\includegraphics[width = 14cm]{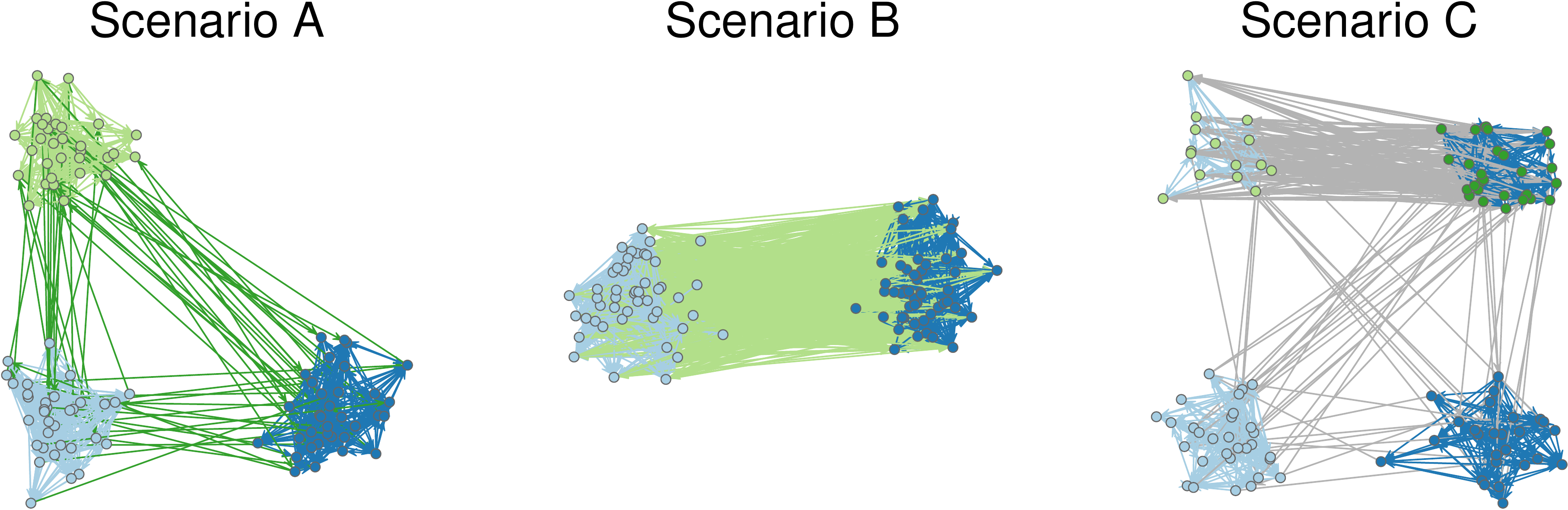}
	\caption{Networks sampled from each scenario. The node colours denote the node cluster memberships and the edge colours denote the majority topic in the corresponding documents.}
	\label{fig:network_examples}
\end{figure}

\begin{itemize}
	\item Scenario $A$ is constituted of three communities, each defining a cluster, and four topics. By definition, a community is a group of nodes more densely connected together than with the rest of the network. For each cluster, a specific topic is employed to sample the documents associated with the intra-cluster connections. Besides, an extra topic is employed to model documents sent between nodes from different clusters. Hence, by construction, the clustering structure can be retrieved either using the network or the texts only.
	\item Scenario $B$ is made of a single community and three topics. Consequently, all nodes connect with the same probability. Then, the nodes are spread into two clusters using distinct topics. An extra topic is used to model documents exchanged between the two clusters. Accordingly, the network itself is not sufficient to find the two clusters but the documents are.
	\item Scenario $C$ comprises three communities and three topics. Two of the communities are associated with their respective topics, say $t_1$ and $t_2$. Furthermore, following Scenario $B$, the third community is split in two clusters, one being associated with topic $t_1$ and the other with $t_2$. Thus, considering both texts and topology, each network is actually made up of four node clusters. Consequently, both textual data and the network are necessary to uncover the clusters. This scenario will be of major interest in this experiment section since it ensures that the two sources of information are correctly used to uncover the node partition.
\end{itemize}
For all scenarios, networks with 100 nodes are sampled and the edges holding the documents are constructed by sampling words from four BBC articles, focusing each on a given topic. The first topic deals with the UK monarchy, the second with cancer treatments, the third with the political landscape in the UK and the last topic deals with astronomy. In the general setting, for all scenarios, the average text length for the documents is set to 150 words.
The parameters used to sample data from the three scenarios are given in Table \ref{tab:summary_scenarii}. Moreover, three examples of networks generated from $A$, $B$ and $C$ are presented in Figure \ref{fig:network_examples}. To summarise, the three proposed scenarios inspect different facets of the model. Scenario $A$ insures that the model rightfully uses the network structure, Scenario $B$ focuses on the usage of the topics to recover the node partition. Finally, Scenario $C$ combines the two scenarios to guarantee that both sources of information are correctly utilised simultaneously.

\begin{table}[h]
	\centering
	\begin{tabular}{|>{\centering}p{0.35\linewidth}|ccc|}
		\hline
		& Scenario $A$ & Scenario $B$ & Scenario $C$\\
		\hline
		$Q$ (clusters) & 3  & 2  & 4 \\
		\hline
		$K$ (topics) & 4 & 3 & 3 \\
		\hline
		Communities & 3 & 1 & 3 \\ 
		\hline
		{$\pi_{qr}$ (connection probabilities)\\ $\psi=0.25$, $\epsilon=0.01$ } & $ 
		\begin{pmatrix}
			\psi & \epsilon  & \epsilon \\
			\epsilon & \psi & \epsilon  \\
			\epsilon & \epsilon & \psi
		\end{pmatrix} $ & 
		$\begin{pmatrix}
			\psi & \psi  \\
			\psi & \psi  
		\end{pmatrix} $ &
		$ 
		\begin{pmatrix}
			\psi & \epsilon  & \epsilon & \epsilon \\
			\epsilon & \psi & \epsilon & \epsilon  \\
			\epsilon & \epsilon & \psi & \psi \\
			\epsilon & \epsilon & \psi & \psi
		\end{pmatrix} $
		\\
		\hline
		Topics matrix $\mathbf{T}$ between pairs of clusters $(q, r)$ & $ 
		\begin{pmatrix}
			t_1 & t_4  & t_4 \\
			t_4 & t_2 & t_4  \\
			t_4 & t_4 & t_3
		\end{pmatrix} $ & 
		$\begin{pmatrix}
			t_1 & t_3  \\
			t_3 & t_2  
		\end{pmatrix} $ &
		$ 
		\begin{pmatrix}
			t_1 & t_3  & t_3 & t_3 \\
			t_3 & t_2 & t_3 & t_3  \\
			t_3 & t_3 & t_1 & t_3 \\
			t_3 & t_3 & t_3 & t_2
		\end{pmatrix} $
		\\ 
		\hline
	\end{tabular}
	\caption{Detail of the three simulation scenarios used to evaluate our model.}
	\label{tab:summary_scenarii}
\end{table}

\paragraph{Clustering performance evaluation} The adjusted rand index (ARI) is used as a measure of the closeness between two partitions. In this paper, ARI compares the true node labels with the node partition provided by a model. In particular, obtaining an ARI of 0 suggests that the clustering is as close to the true node labels as a random cluster assignment of the nodes. On the contrary, the closer the ARI is to 1, the better the results are. Ultimately, an ARI of 1 signifies that the true partition was perfectly recovered (up to a label permutation).

\paragraph{Level of difficulty} In order to generate more situations from the three scenarios, we introduce the \textit{Hard} difficulty to test the model robustness against two aspects. First, we want to test the model against documents using several topics. Thus, in the \textit{Hard} difficulty, the documents are formed of multiple topics such that, for any edge $(i,j)$ with node $i$ in cluster $q$ and node $j$ in cluster $r$, the topic proportions are computed as a ratio between the pure topic proportions $\tilde{Y}_{qr}^{\star} \in \{0,1\}^{K}$, with zeros everywhere except at the coordinate corresponding to the true topic, and between the uniform distribution over the topics. This combination is controlled by a parameter $\zeta$ such that $\zeta=0$ corresponds to a pure topic case while $\zeta=1$ leads to a uniform distribution over the topics. This translates into:
\begin{align}
	\tilde{Y}_{qr} = (1-\zeta) \tilde{Y}_{qr}^{\star} + \zeta * \left(\frac{1}{K}, \dots, \frac{1}{K}\right)^{\top},
\end{align}
with $\zeta = 0.7$ in the \textit{Hard} setting. The second aspect tested by the \textit{Hard} setting is the robustness in the presence of less connected communities. Consequently, the intra-cluster connection probability is decreased from $\psi = 0.25$ in the classical setting to $\psi = 0.1$ in the \textit{Hard} one.

\subsection{Main features of Deep-LPTM}\label{sec:main_features_deep_lptm}
This section gives an overview of the main features of Deep-LPTM on one network simulated according to Scenario $A$, with an intra cluster connection probability $\psi$ equal to $0.15$, an inter-cluster connection probability $\epsilon$ fixed to $0.05$ and the parameter controlling the topic proportions  $\zeta$ set to $0.5$. In addition, $\phi_Y, \rho, \alpha$, the parameters referring to the topic modelling, are pre-trained for only 5 epochs with ETM alone. Conversely, the parameters $\phi_Z, \kappa, \mu, \sigma$, related to network modelling, are randomly initialised without pre-training to illustrate the evolution of the node embeddings during the optimisation. In the rest of the paper, those parameters will be pre-trained by running ETM and Deep-LPM independently before hand. 

\begin{figure}
\centering
\includegraphics[width=0.5\textwidth]{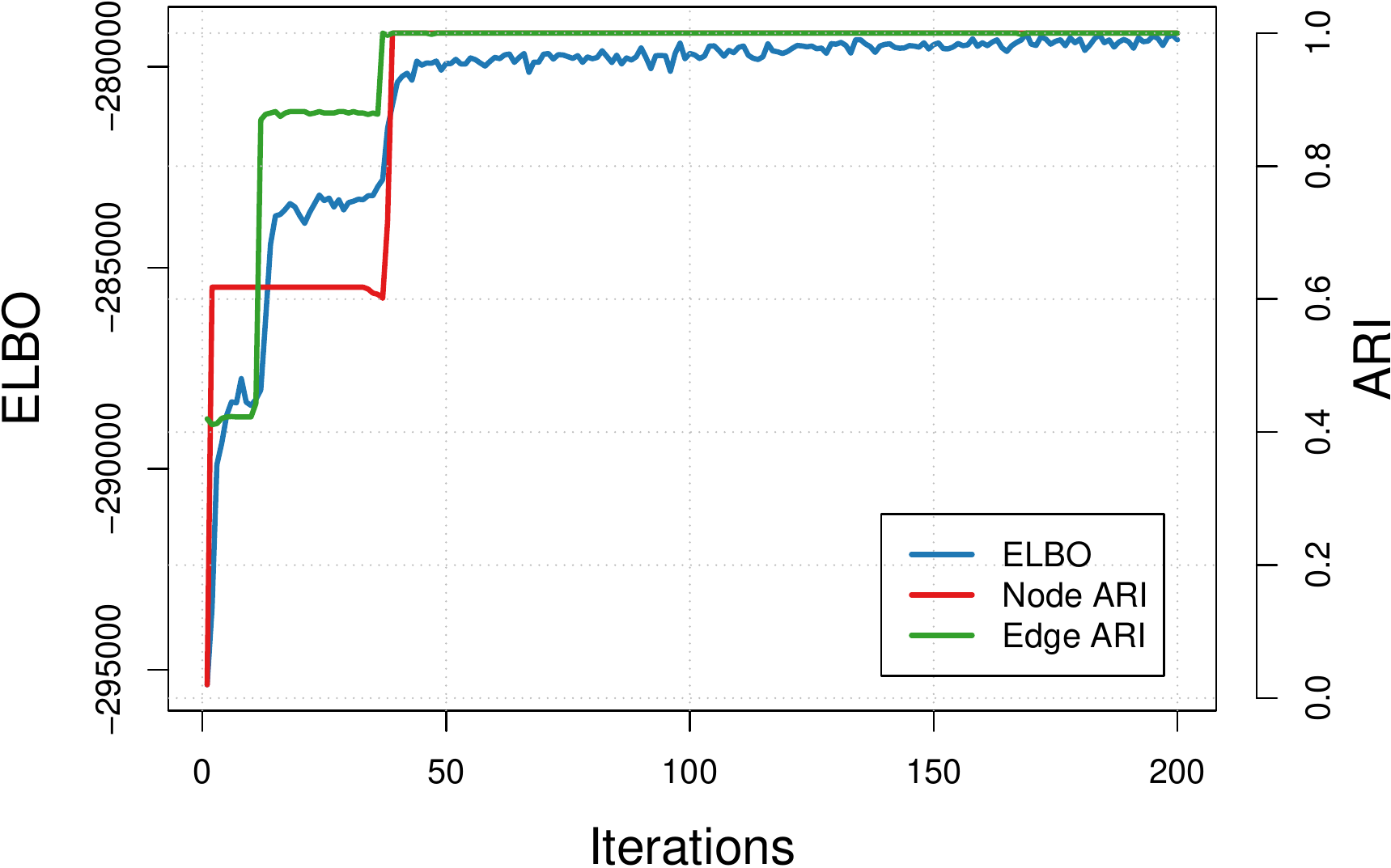}
\captionof{figure}{\normalsize Evolution of the ELBO, as well as the node and edge ARI during the optimisation of Deep-LPTM.}
\label{fig:example_sc_a_elbo_aris}
\end{figure}

On the one hand, the evolution of the ELBO as well as the ARI of the nodes and the edges are presented in Figure \ref{fig:example_sc_a_elbo_aris}. The node ARI and the edge ARI increase to reach an ARI of $1$ following the evolution of the ELBO. We only display the first 200 epochs for the sake of clarity, but the entire training is provided in the appendix. 

\begin{figure}
	\centering
	\includegraphics[width=17cm]{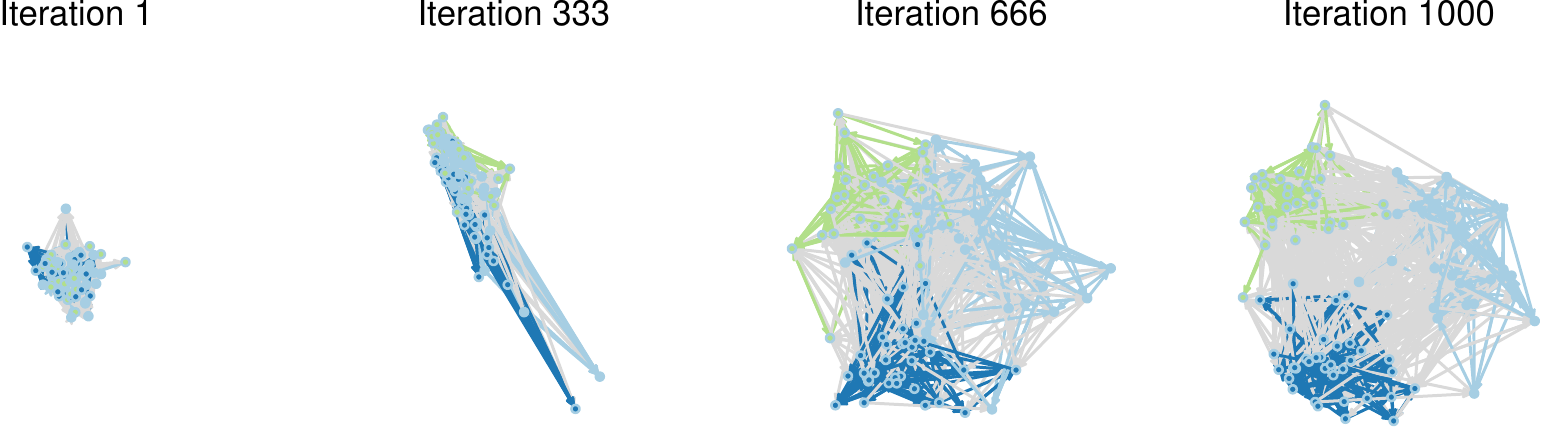}
	\caption{Evolution of the latent node positions during the training of Deep-LPTM. \hspace*{10cm}}
	\label{fig:example_sc_a_embeddings_evolution}
\end{figure}

On the other hand, Figure \ref{fig:example_sc_a_embeddings_evolution} features the evolution of the node latent positions during the training. Interestingly enough, Deep-LPTM finds a meaningful representation of the network even with a random initialisation. This difficult problem requires to train the model longer when no initialisation is provided. Hence, the ELBO continues to increase as displayed in Figure \ref{fig:example_sc_a_elbo_aris_1000}, presented in the appendix. In the rest of the paper, the GCN parameters as well as the topic model parameters are initialised before hand. In addition, the node cluster memberships probabilities are initialised with a similarity-based method  between the topic proportions $\tilde{Y}$ of the node neighbours \cite{bouveyron2018stochastic}.

\vspace*{0.5cm}
\begin{minipage}[c]{0.35 \linewidth}
		\includegraphics[width=\linewidth]{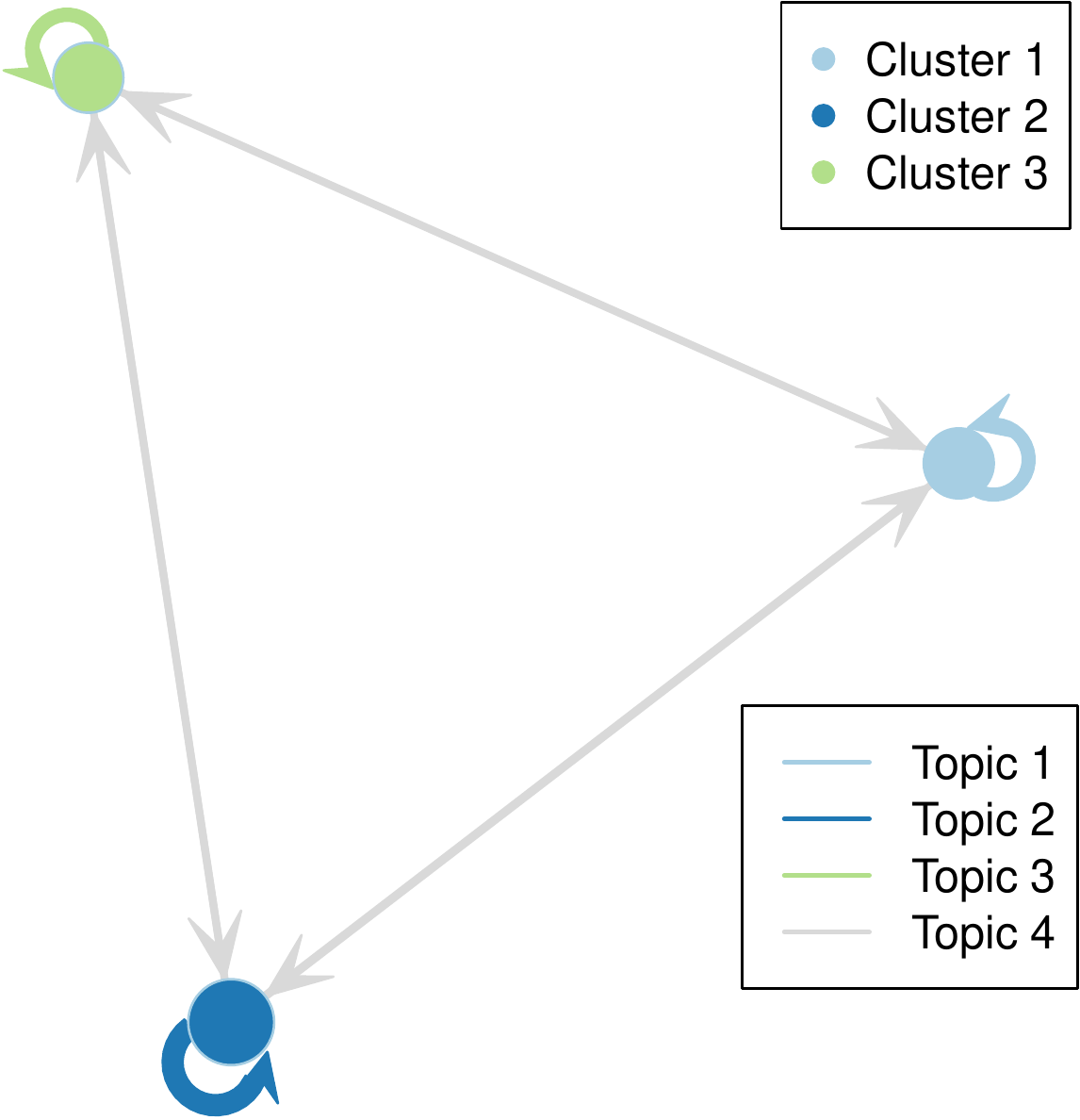}
\end{minipage}
\hfill
\begin{minipage}[c]{0.6 \linewidth}
		\captionof{figure}{The meta-network is a representation of the network at the cluster level  based on Deep-LPTM estimates. Each cluster is represented by a node, the node sizes depend on the number of nodes assigned to each cluster, the node positions as well as the major topics between two connected clusters (denoted by the colours of the edges) are estimated by the parameters $(\mu_q)_q$ as well as $(m_{qr})_{qr}$ respectively and the sizes of the edges depend on the number of connections between two clusters.}
\label{fig:meta_network_example}
\end{minipage}

\begin{table}[ht]
	\centering
	\begin{tabular}{rllll}
		\toprule
		{} &          1 &         2 &              3 &           4 \\
		\midrule
		1 &     cancer &     black &       princess &       seats \\
		2 &       cell &      hole &          birth &       david \\
		3 &      occur &   gravity &      charlotte &   political \\
		4 &      genes &     light &      cambridge &        lost \\
		5 &    cancers &    shadow &          queen &     kingdom \\
		6 &        due &    credit &  granddaughter &       black \\
		7 &  mutations &     event &        duchess &       party \\
		8 &  radiation &      disc &         palace &        part \\
		9 &   princess &  princess &         london &  resentment \\
		10 &    include &   horizon &          great &      united \\
		\bottomrule
	\end{tabular}
	\captionof{table}{Topics of the model in Scenario A Easy, represented by the 10 most probable words per topic. \hspace*{10cm}}
	\label{tab:topic_sc_A_easy}
\end{table}
\vspace*{-0.7cm}
The meta-network, represented in Figure \ref{fig:meta_network_example}, describes the connectivity at the cluster level. The node sizes depend on the number of nodes assigned to each cluster $q$ and are given by $\sum_{i=1}^N \hat{C}_{iq}$, with $\hat{C}_{iq} = 1$ if $\argmax \tau_{i} = q$ and $0$ otherwise. The cluster positions are estimated by $(\mu_q)_q$ and the major topics (denoted by the colours of the edges on the figure) between two clusters is estimated by the $\argmax m_{qr}$  for all pairs of clusters $(q,r)$. Finally, the sizes of the edges depend on the number of connections between clusters, given by $ \sum_{i,j =1}^N \hat{C}_{iq} \hat{C}_{jr}$ for all pairs of clusters $(q,r)$.

Eventually, Table \ref{tab:topic_sc_A_easy} presents the topics obtained by Deep-LPTM. They are both very interpretable and distinguishable one from another which is crucial to understand complicated datasets. This will be stressed in the analysis of Enron email dataset in Section \ref{sec:real_world_dataset}.

\subsection[Impact of initialisation]{Impact of the initialisation and the pre-trained embeddings} \label{sec:initialisation_impact}

This section aims at evaluating the improvement of our method upon the initialisation, with a warm start and without. In this regard, Table \ref{tab:comparison_init_and_results} presents the ARI of a random initialisation as well as a dissimilarity initialisation \cite{bouveyron2018stochastic} ,denoted \textit{random} and \textit{dissimilarity} respectively, in Table \ref{tab:comparison_init_and_results}. The initialisation alone (without any model name preceding it in the table)  as well as the model with the initialisations (with the model preceding the initialisation) are provided. Moreover, the Deep-LPTM node clustering is evaluated with and without pre-trained skipgram embeddings \cite{mikolov2013efficient}, denoted \textit{PT} in the table. The results are obtained by averaging the ARI over 10 graphs for each scenario and difficulty and can be summarised in three points.

First, in all cases where the initialisation has not already reached an ARI of $1$, Deep-LPTM improves the node clustering, even in difficult settings with no warm-start. For instance, in Scenario $A$ with the \textit{Hard} setting, the model starts from an ARI of $0.31$, with the dissimilarity initialisation, to reach $0.99$ and $1.00$ without and with pre-trained embeddings respectively. 

Second, the improvement provided by the pre-trained embeddings depends on the scenario. On the one hand, Scenario $A$ benefits from the usage of pre-trained embeddings which always improves the results. For instance, in the \textit{Hard} setting with a random initialisation, the ARI increases from $0.80 \pm 21$ to $0.95 \pm 0.05$. On the other hand, Scenario $B$ and $C$ always favour the results without pre-trained embeddings. As an example, with a random initialisation in the \textit{Hard} setting, the ARI decreases from $0.95 \pm 0.05$ to $0.73 \pm 0.30$ and $0.47 \pm 0.02$ to $0.45 \pm 0.04$ in Scenario $B$ and $C$ respectively. The same deduction can be made with the dissimilarity initialisation. Since Scenario $B$ and $C$ are the ones evaluating the text contribution to the clustering, we advise not to use pre-trained embeddings.

Finally, the best ARI are all obtained with the dissimilarity initialisation. Consequently, Deep-LPTM will only be initialised with it in the rest of the paper.

\begin{table}
	\centering
	\begin{tabular}{llccc}
		\toprule
		&           &         ScenarioA &         ScenarioB &         ScenarioC \\
		&           &          Node ARI &          Node ARI &          Node ARI \\
		\midrule
		\multirow{5}{*}{Easy} & Random init &   0.00 $\pm$ 0.00 &   0.00 $\pm$ 0.00 &   0.00 $\pm$ 0.00 \\
		& Dissimilarity init&   0.97 $\pm$ 0.06 &   1.00 $\pm$ 0.00 &   0.98 $\pm$ 0.03 \\
		\cline{2-5}
		& Deep-LPTM  random &   1.00 $\pm$ 0.00 &   1.00 $\pm$ 0.01 &   0.63 $\pm$ 0.20 \\
		& Deep-LPTM dissim&   1.00 $\pm$ 0.00 &   1.00 $\pm$ 0.00 &   1.00 $\pm$ 0.00 \\
		& Deep-LPTM - PT random &   1.00 $\pm$ 0.00 &   0.90 $\pm$ 0.30 &   0.55 $\pm$ 0.15\\
		& Deep-LPTM - PT dissim &   1.00 $\pm$ 0.00 &   1.00 $\pm$ 0.00 &   1.00 $\pm$ 0.00 \\
		\midrule
		\multirow{5}{*}{Hard}& Random init &   0.00 $\pm$ 0.00 &   0.00 $\pm$ 0.00 &   0.00 $\pm$ 0.00\\
		& Dissimilarity init&   0.31 $\pm$ 0.14 &   1.00 $\pm$ 0.00 &   0.38 $\pm$ 0.24 \\
		\cline{2-5}
		&Deep-LPTM random &   0.80 $\pm$ 0.21 &   0.95 $\pm$ 0.05 &   0.47 $\pm$ 0.02  \\
		& Deep-LPTM dissim &   0.99 $\pm$ 0.02 &   1.00 $\pm$ 0.00 &   0.89 $\pm$ 0.15 \\
		& Deep-LPTM - PT random &   0.95 $\pm$ 0.05 &   0.73 $\pm$ 0.30 &   0.45 $\pm$ 0.04 \\
		& Deep-LPTM - PT dissim &   1.00 $\pm$ 0.01 &   1.00 $\pm$ 0.00 &   0.85 $\pm$ 0.18 \\
		\bottomrule
	\end{tabular}
	\caption{ Adjusted rand index (ARI) of the initialisations and the results of Deep-LPTM in terms of node clustering, without and with pre-trained embeddings (denoted PT in that case). ARI is averaged over 10 graphs, for each scenario and difficulty.}
	\label{tab:comparison_init_and_results}	
\end{table}

\subsection{Model selection}\label{sec:model_selection_xp}
In order to assess the model selection criterion, this section provides two experiments. First, we evaluate IC2L relevancy to select $Q$ on all three scenarios. Second, we test IC2L efficiency to select the triplet $(K,P,Q)$ on Scenario $C$ specifically.

\paragraph{Selection of $Q$ with $P=2$ and the true $K$}
Keeping $P$ set to $2$ and $K$ fixed to its true value for the moment, Table \ref{tab:benchmark} assesses the effectiveness of IC2L to select the number of clusters $Q$. In all three scenarios, IC2L selects the true model 10 times out of 10.

\begin{table}
	\centering
	\begin{tabular}{lccc}
		\toprule
		{} & Scenario A & Scenario B & Scenario C \\
		{} & $Q^{\star} = 3$ & $Q^{\star} = 2$ &  $Q^{\star} = 4$ \\
		\midrule
		$Q = 2$  &      $0$ &        $10$ &         $0$ \\
		$Q = 3$  &     $10$ &         $0$ &         $0$ \\
		$Q = 4$ &      $0$ &         $0$&         $10$ \\
		$Q = 5$  &      $0$ &        $0$ &        $0$\\
		$Q = 10$ &      $0$ &         $0$ &       $0$ \\
		\bottomrule
	\end{tabular}
	\caption{Number of times a value $Q$ is selected by the IC2L criterion over 10 graphs with the true value of $K$ and $P = 2$. \hspace*{10cm}}
	\label{tab:benchmark}
\end{table}

\paragraph{Selection of the triplet $(P,K,Q)$}
Let us now consider selecting the triplet $(K,P,Q)$ simultaneously. Table \ref{tab:benchmark_scenario_C} displays the number of times a triplet is selected by IC2L for 10 graphs simulated according to Scenario $C$ (with $Q^{\star} = 4$ and $K^{\star} = 3$ the true values). The selected node embedding dimension is always $P=2$, thus, we only provide $K$ and $Q$ in the table. First, it is satisfactory that IC2L always selects the true number of topics and clusters. Second, by always picking the dimension $P=2$, IC2L favours models with a lower complexity. In our case, this translates into choosing models able to directly visualise the data in two dimensions, simply by plotting the latent vectors $Z_i$. This suits our purpose of building an explainable model.   

\begin{table}
	\centering
	\begin{tabular}{lccccc}
		\toprule
		{} &  $K=2$ &  $\bf{K=3}$ &  $K=4$ &  $K=5$ &  $K=6$ \\
		\midrule
		$Q=2$ &    $0$ &    $0$ &    $0$ &    $0$ &    $0$ \\
		$Q=3$ &    $0$ &    $0$ &    $0$ &    $0$ &    $0$ \\
		$\bf{Q=4}$ &    $0$ &    $10$ &      $0$ &    $0$ &    $0$ \\
		$Q=5$ &    $0$ &    $0$ &    $0$ &    $0$ &    $0$ \\
		$Q=6$ &    $0$&     $0$ &    $0$ &    $0$ &    $0$ \\
		\bottomrule
	\end{tabular}
	\caption{Number of times a triplet $(K,P,Q)$ is associated with the highest IC2L over 10 graphs simulated according to Scenario C ($Q^{\star} = 4$ and $K^{\star} = 3$). All the models with the highest IC2L value correspond to  $P=2$. Therefore, only the table corresponding to this value is shown.}
	\label{tab:benchmark_scenario_C}
\end{table}

\subsection{Benchmark}
\begin{table}[ht]
	\centering
	\begin{tabular}{llccc}
		\toprule
		&      &         ScenarioA &         ScenarioB &         ScenarioC \\
		\midrule
		\multirow{5}{*}{Easy} 
		& SBM &    1.00 $\pm$ 0.00 &   -0.00 $\pm$ 0.01 &   0.73 $\pm$ 0.05 \\
		& STBM &   1.00 $\pm$ 0.00 &   1.00 $\pm$ 0.00 &   1.00 $\pm$ 0.01 \\
		& ETSBM &   0.99 $\pm$ 0.03 &   1.00 $\pm$ 0.00 &   0.96 $\pm$ 0.04 \\
		& ETSBM - PT &   1.00 $\pm$ 0.00 &   1.00 $\pm$ 0.00 &   0.96 $\pm$ 0.05 \\
		\cmidrule{2-5}
		& Deep-LPTM &   1.00 $\pm$ 0.00 &   1.00 $\pm$ 0.00 &   1.00 $\pm$ 0.00 \\
		& Deep-LPTM - PT &   1.00 $\pm$ 0.00 &   1.00 $\pm$ 0.00 &   1.00 $\pm$ 0.00 \\
		\cmidrule{1-5}
		\multirow{5}{*}{Hard} 
		& SBM &    0.97 $\pm$ 0.03 &   0.00 $\pm$ 0.00 &   0.62 $\pm$ 0.1 \\
		& STBM &   0.63 $\pm$ 0.23 &   1.00 $\pm$ 0.00 &   0.66 $\pm$ 0.19 \\
		& ETSBM &   0.96 $\pm$ 0.10 &   0.90 $\pm$ 0.30 &   0.72 $\pm$ 0.25 \\
		& ETSBM - PT &   0.99 $\pm$ 0.01 &   1.00 $\pm$ 0.00 &   0.74 $\pm$ 0.21 \\
		\cmidrule{2-5}
		& Deep-LPTM &   0.99 $\pm$ 0.02 &   1.00 $\pm$ 0.00 &   0.89 $\pm$ 0.15 \\
		& Deep-LPTM - PT &   1.00 $\pm$ 0.01 &   1.00 $\pm$ 0.00 &   0.85 $\pm$ 0.18 \\
		\bottomrule
	\end{tabular}
	\caption{ARI of the node clustering averaged over 10 graphs in all three scenarios for the two levels of difficulty Easy and Hard. Deep-LPTM, as well as ETSBM, are presented with and without pre-trained embeddings (denoted PT). Moreover, STBM and SBM are also provided as baselines.}
	\label{tab:full_benchmark}
\end{table}

We close this section with a benchmark study presented in Table \ref{tab:full_benchmark} to compare Deep-LPTM with state-of-the-art ETSBM and STBM. We also provide SBM and Deep-LPM as baselines even though they are not able to take into account the text edges. As in the previous sections, the table presents the average of the ARI over 10 graphs. Each graph result is obtained by running each method with five different initialisations and by taking the one resulting in the highest ELBO.
The table is presented for three different models, namely STBM, ETSBM and Deep-LPTM. The last two models are evaluated with and without pre-trained embedding.  In all cases, Deep-LPTM is either as good as or better than other models. In particular, in Scenario $C$ with difficulty \textit{Hard}, the ARI of Deep-LPTM node clustering is higher than all other methods, by at least $0.15$. Likewise, in Scenario $A$ with difficulty \textit{Hard}, Deep-LPTM always recover the true partition while STBM only reaches an ARI of $0.66 \pm 0.18$. 

\section{Application to the analysis of the Enron email network}\label{sec:real_world_dataset}

\begin{figure}
	\centering
	\includegraphics[width=0.9\linewidth]{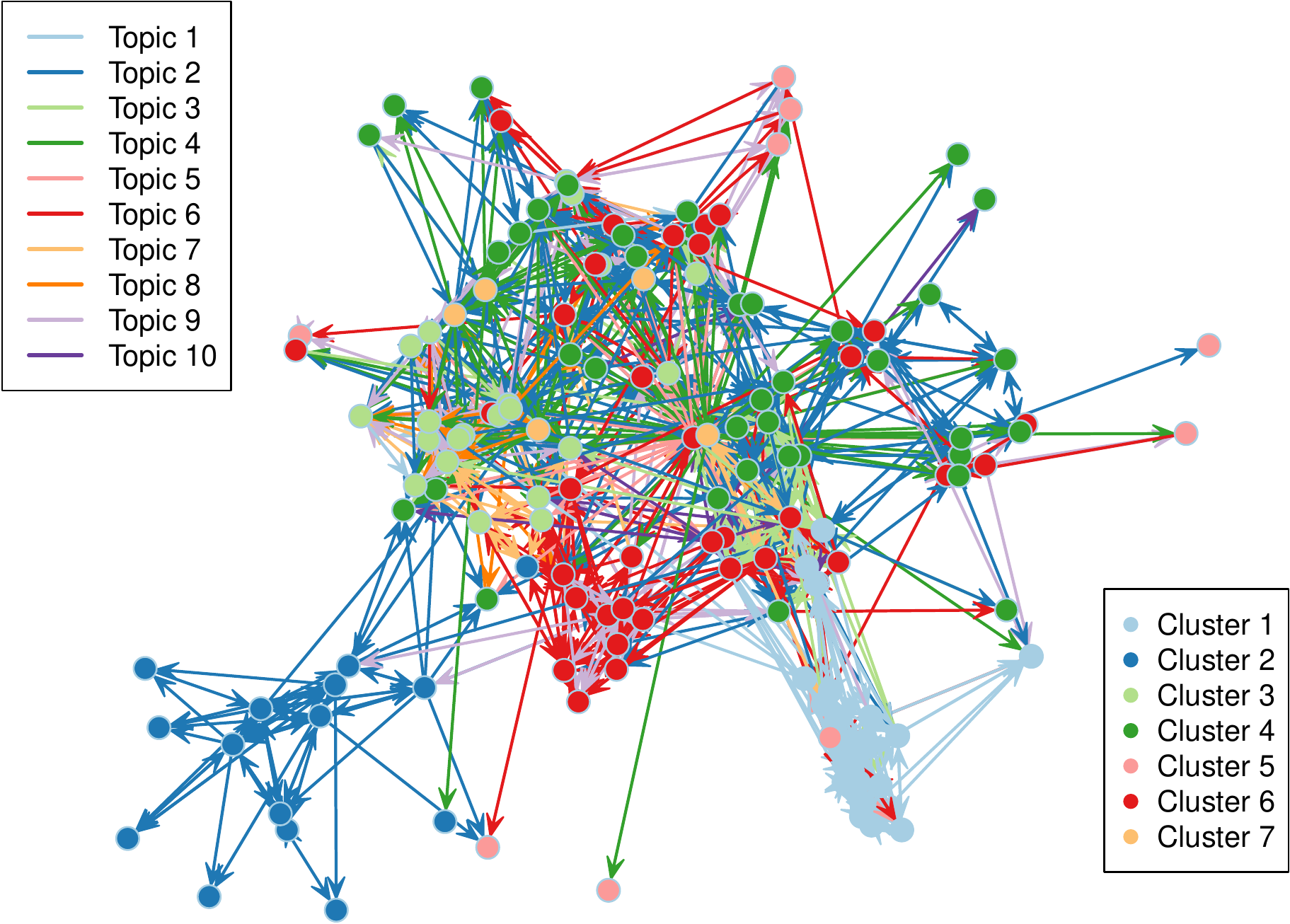}
	\caption[caption]{Deep-LPTM representation of Enron email network. The nodes positions, the node cluster memberships (denoted by the colour of the nodes) as well as the majority topic in the documents (denoted by the colour of the edges) are estimated by Deep-LPTM.}
	\label{fig:ENRON_network}
\end{figure}

\begin{figure}[t]
	\centering
	\captionsetup[subfloat]{captionskip=0pt}
	\subfloat{\includegraphics[width=11cm]{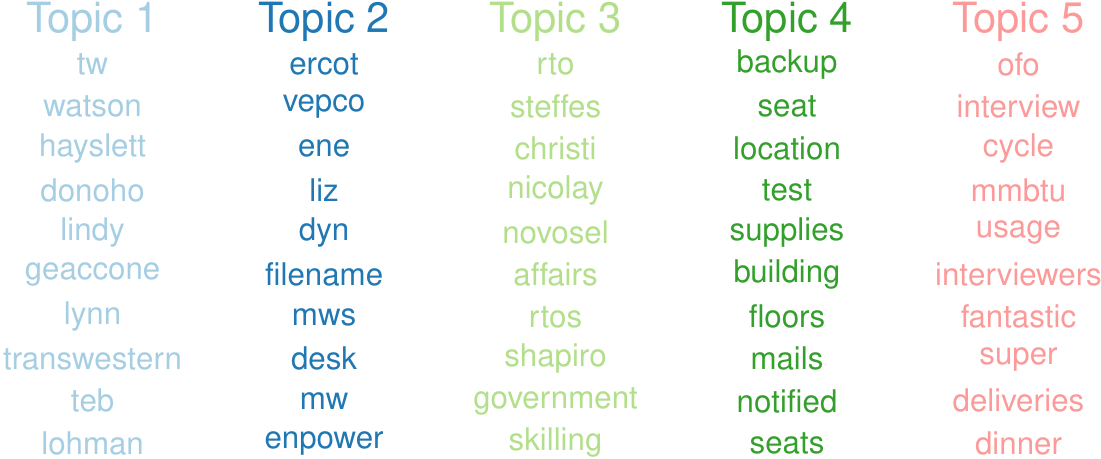}}\\
	\subfloat{\includegraphics[width=11cm]{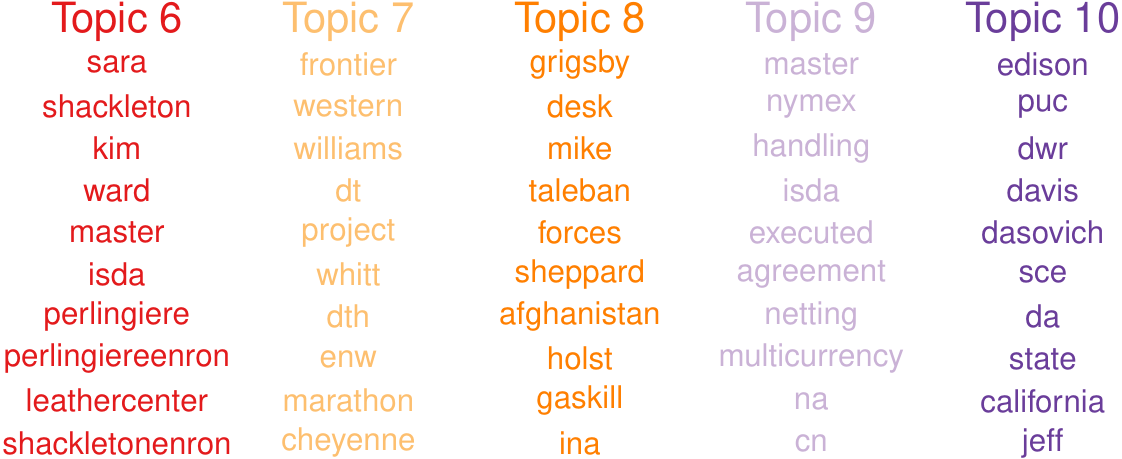}}
	\caption{{The 10 most probable words of each topic according to Deep-LPTM.} \hspace*{10cm}}
	\label{fig:ENRON_topics}
\end{figure}

Enron, formed in 1985, was an American company selling natural gas in North America. In 2001, the securities and exchange commission (SEC) opened an investigation on October, 31th, for fraud, while on August, 14th, the company was "probably in the strongest and best shape that it [had] probably ever been in" according to its CEO. On early December the same year, the company filed for the largest bankruptcy at that time. We propose here to concentrate on the critical period September, 1st to December, 31th leading to the downfall of the company in order to understand the organisation of the company during this crucial period. Thanks to the decision of the federal energy regulatory commission (FERC), the dataset is publicly accessible, and contains all 20,940 emails exchanged between 149 employees. All edges holding multiple messages were coerced into a single meta-message by stacking the documents together. As a result, the network holds 1,234 edges between the 149 employees. The dataset can be found at \url{https://www.cs.cmu.edu/~./enron/}.

The estimation of Deep-LPTM is conducted for all triplets $(Q,K,P)$ where $Q \in \{5,7,10\}$, $K \in \{3, 5, 7, 10\}$ and $P \in \{2,4,8,16\}$. The highest value of IC2L corresponds to the triplet $(Q,K,P) = (7, 10, 2)$.  Deep-LPTM clustering results are displayed in Figure \ref{fig:ENRON_network}. The node cluster memberships as well as the edge majority topics are represented by their respective colours. In addition, the topics are interpreted by looking at their corresponding most probable words in Figure \ref{fig:ENRON_topics}. Interestingly, IC2L selected a low-dimensional node latent space, which fits with the observations made in \cite{hoff2002latent}.

\paragraph{Topics analysis} The topics can be depicted as follow:
\begin{itemize}
	\item Topic 1 concerns Charles Watson, Dynegy CEO at the time, that negotiated a deal to finance Enron, involving the \textit{transwestern} pipeline, and to merge the companies
	
	\item Topic 2 refers to regional energy 
	
	\item Topic 3 deals with business operations
	
	\item Topic 4 is related to office supplies and day-to-day work
	
	\item Topic 5 mentions the energy usage and delivery 
	
	\item Topic 6 is related to legal and strategical aspects of Enron business, involving Sara Shakleton (vice president of Enron North America Corporation), and Debra Perlingiere, from the legal department 
	
	\item Topic 7 is concerned with infrastructures and geographical projects
	
	\item Topic 8 corresponds to a discussions about Enron activities in Afghanistan, which may be seen as sensitive given the american situation in 2001
	
	\item Topic 9 focuses on financial aspects
	
	\item Topic 10 mentions the California electricity crisis, which almost led to the bankruptcy of the Southern California Edison corporation
	
\end{itemize}
The topics as well as the visualisation provide significant information on the dataset. In particular, Deep-LPTM identifies different departments and cases of the company through the topics and successfully represent it in the graph structure as we shall detail.

\begin{figure}
	\centering
	\includegraphics[width=0.6\linewidth]{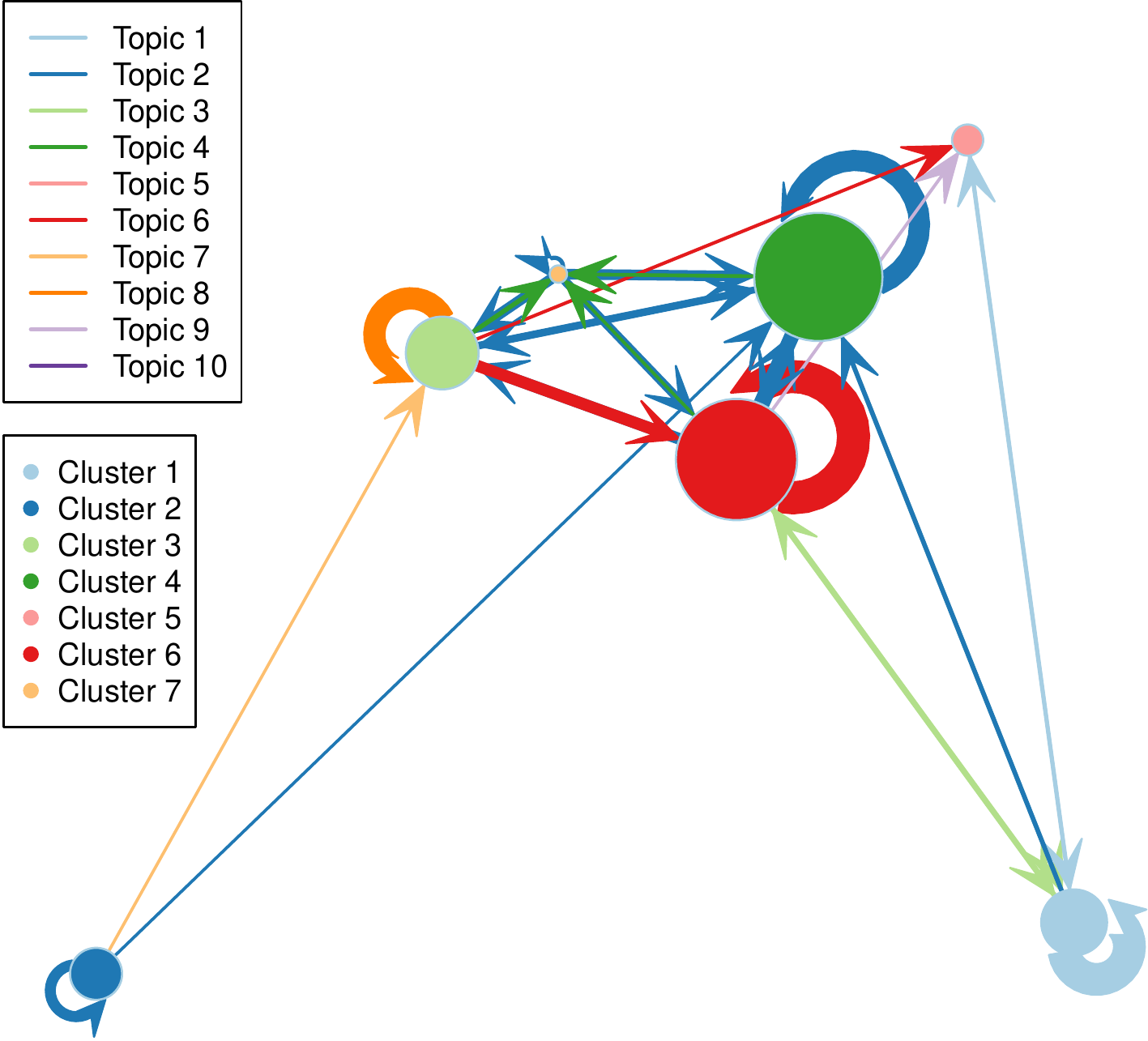}
	\caption{Enron email meta-network, based on Deep-LPTM estimates, represents the clusters and their interactions. The node sizes depend on the number of email accounts assigned to each cluster and the node positions are estimated by Deep-LPTM. The colours of the edges refer to the majority topic between the respective clusters and the sizes of the edges depend on the number of connections between the clusters. We keep only the edges corresponding to more than five connections for readability purposes.}
	\label{fig:meta_network_enron}
\end{figure}

\paragraph{Cluster analysis}
The meta-network in Figure \ref{fig:meta_network_enron} gives an overview of the results to analyse the clusters. We keep only the edges corresponding to more than five connections for readability purposes. However, important details provided by Figure \ref{fig:ENRON_network} could not be uncovered with the meta-network only.

First, Clusters 1 and 2 are well separated from the rest of the graph. Moreover, each one is characterised by a specific topic. Cluster 1 is involved in discussions about a financial deal and a possible merge (Topic 1) and more than half of the employees in Cluster 1 have a managerial position or work in the legal department (vice presidents, directors, lawyer). Cluster 2 refers to the regional energy business (Topic 2) and nine out of the 15 nodes correspond to either employees or a traders.

Second, Cluster 6 displays a high level of internal connectivity and is essentially featured by discussions involving the legal department (Topic 6) as well as financial aspects of the company (Topic 9). The status of the employees in that cluster are mainly composed of directors, vice presidents as well as presidents of the company. Interestingly, Deep-LPTM placed a president, corresponding to the node with the highest degree in the graph, at the centre of the network, and allocated it to the cluster of managers of the company. This graphical property, unique to Deep-LPTM, stresses the incorporation of the connections, as well as the topics in the emails, to obtain a meaningful representation of the network. Conversely, in Figure \ref{fig:meta_network_enron}, the meta-network is not able to dissociate the connectivity of a node with a different connectivity than the rest of the cluster. Thus, Cluster 6 is central as a whole in the network in Figure \ref{fig:meta_network_enron}.

Third, the topics involved in emails of Clusters 3 and 4 are the main drivers of the characterisation of the two clusters. For instance, Cluster 3 is highly connected to the graph and is involved in discussions about infrastructures and geographical aspects (Topic 7), as well as financial aspects (Topic 9). Emails related to Afghanistan were exchanged, mainly involving nodes from Cluster 3. It is worth noticing that nodes in cluster 3 correspond to people with a high position in the hierarchy of the company (almost two third of the employees in the clusters are either managers, directors, vice presidents, presidents or CEO). In addition, although many nodes in Cluster 4 are involved in regional energy discussions (Topic 2) like Cluster 2, they also discuss about day to day business questions (Topic 4), and are highly connected with the rest of the network, unlike Cluster 2. 

Finally, Cluster 7 is composed of four nodes that are at the crossroads between several clusters and topics. We close this analysis with Cluster 5 that is composed of nodes mainly receipting emails and poorly connected to the graph, highlighted in the representation by the node positions being on the rim of the network.


\section{Conclusion}
We introduced a novel deep probabilistic model, named the deep latent position topic model (Deep-LPTM) to analyse networks with textual edges. Deep-LPTM allows to simultaneously cluster nodes, to model the topics in the documents, and to provide a network visualisation. To represent networks into a Euclidean space, most methodologies rely either on heuristic method used a posteriori or only focus on the connectivity of the graph. On the contrary, this work tackles this problem by incorporating the network representation into the modelling, enabling the clustering as well as the topic modelling to be included in the calculation of the node positions. To benefit from the flexibility of deep neural networks, our methodology is based on a graph convolutional network (GCN) to encode the nodes into a vector space using the connectivity of the graph as well as a neural topic model to encode the documents and the topics into a vector space. Even though we focused on directed networks, the extension to undirected networks is straightforward. The applications of this methodology are numerous, including social sciences, journalism or social network analysis. The proposed methodology relies on a variational inference algorithm to maximise the marginal likelihood. The optimisation combines analytical formulas and stochastic gradient descent steps to estimate the parameters. In this paper, we also derived the integrated classification and latent likelihood (IC2L) criterion to choose relevant numbers of clusters and topics as well as the dimension of the node latent space. Both the extensive benchmark study as well as the Enron emails analysis highlighted the visualisation power and the clustering efficiency of the proposed methodology. In future work, it is crucial to adapt this methodology to networks with multiple edges, corresponding to the exchange of several documents between nodes.

\appendix

\section{Graphical Model}
\begin{figure}[h]
	\centering	
	\includegraphics[width=0.8\linewidth]{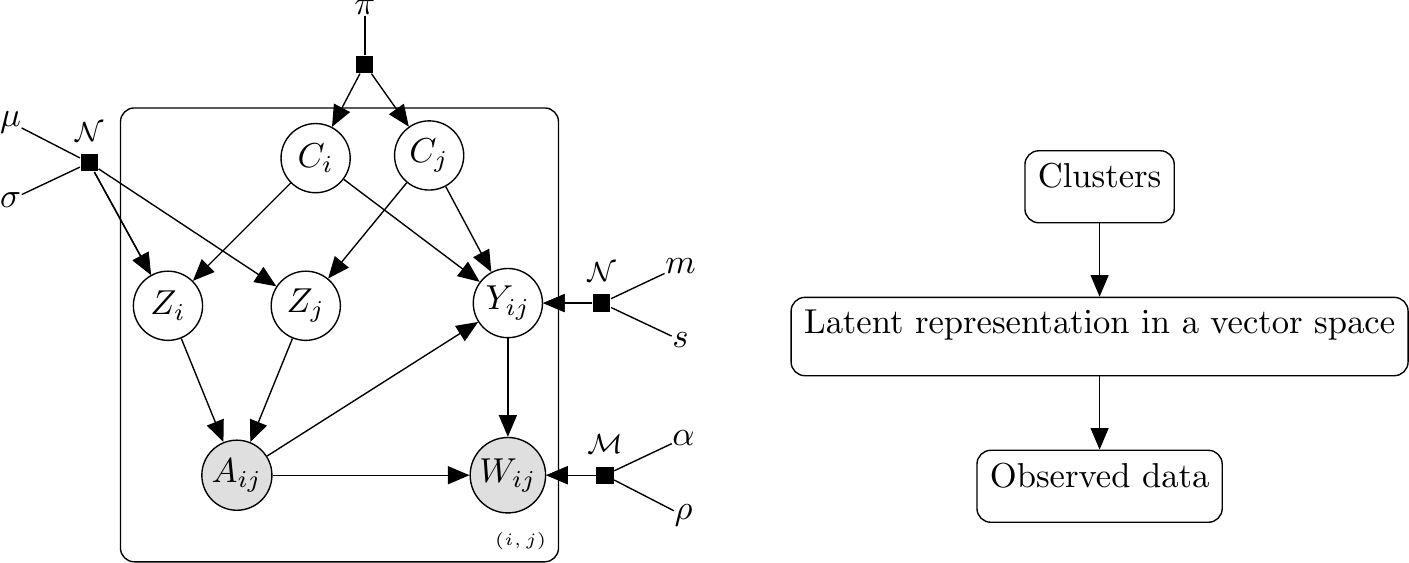}
	\caption{Graphical model with the parameters of Deep-LPTM where the $Z_i$s denote the latent node vectors, $Y_{ij}$s the latent document vector, $C_i$s the node cluster memberships, $A = (A_{ij})_{ij} \in \mathcal{M}_{N \times N} (\{0,1\})$ the binary adjacency matrix and $W_{ij}$ the document sent by node $i$ to node $j$.}
	\label{fig:graphical_model_full}
\end{figure}

Figure \ref{fig:graphical_model_full} provides the graphical representation of the model with its parameters.

\section{Computation of the ELBO terms}\label{sec:elbo_computation_details}
In this section, computational details regarding the ELBO are provided term by term. First, $\mathbb{E}_{R}\left[ \log p(A \mid Z, \kappa) \right]$  is given by:
\begin{align*}
	\mathbb{E}_{R}\left[ \log p(A \mid Z, \kappa) \right] = \sum_{i,j = 1}^N \bigl\{ A_{ij}\mathbb{E}_{R}\left[\log p_{ij}\right] + (1 - A_{ij})\mathbb{E}_{R}\left[\log \bigl( 1- p_{ij} \bigr) \right]\bigr\},
\end{align*}
where  $p_{ij} = \left(1 + e^{- \eta_{ij}} \right)^{-1}$ and $\eta_{ij} := \kappa - \| Z_i - Z_j \|$.

Second,  denoting $\beta_k = \softmax(\rho^{\top} \alpha_k) \in \mathbb{R}^{V}$, $\beta = (\beta_1 \dots \beta_K)^{\top} \in \mathcal{M}_{K \times V}(\mathbb{R})$ and  $w_{ij}^v = (\beta^{\top} \tilde{Y}_{ij})_{v}$, the probability for the word $v$ to appear in document $W_{ij}$ ,for any $v \in \{1,\dots,V\}$,  $\mathbb{E}_{R}\left[ \log p(W \mid A, Y, \rho, \alpha) \right]$ is:
\begin{align*}
	\mathbb{E}_{R}\left[ \log p(W \mid A, Y, \rho, \alpha) \right] & = \sum_{i,j = 1}^{N} A_{ij}\mathbb{E}_{R}\left[ \log \mathcal{M}_{V}\left(W_{ij}; M_{ij}, \beta^{\top} \softmax(Y_{ij}) \right) \right] \nonumber\\
	& = \sum_{i,j} A_{ij}\mathbb{E}_{R}\left[ \log \frac{ M_{ij}!}{\prod_{v=1}^V (W_{ij}^v)!} \prod_{v=1}^V (w_{ij}^v)^{W_{ij}^v} \right].
\end{align*}
The difference between the terms related to the cluster memberships, $\mathbb{E}_{R}\left[ \log p(C \mid \pi) \right]$ and $\mathbb{E}_{R}\left[\log R(C) \right]$, gives the following:
\begin{align*}
	\mathbb{E}_{R}\left[ \log p(C \mid \pi) \right] -\mathbb{E}_{R}\left[\log R(C) \right] & = \sum_{i=1}^M \sum_{q=1}^Q \mathbb{E}_{R}\left[ C_{iq} \log \pi_{q} \right] -\mathbb{E}_{R}\left[ C_{iq} \log \tau_{iq} \right]\nonumber\\
	& = \sum_{i=1}^N \sum_{q=1}^Q \tau_{iq} \log \frac{\pi_{q}}{\tau_{iq}}.
\end{align*}
The difference between the generative distribution of the node positions term $\mathbb{E}_{R}\left[ \log p(Z \mid C, \mu, \sigma) \right]$ and the one related to the variational distribution of the node positions  $\mathbb{E}_{R}\left[ \log R(Z \mid A) \right] $ gives:
\begin{align}\label{eq:KL_nodes}
	&\mathbb{E}_{R}\left[ \log p(Z \mid C, \mu, \sigma)\right] -\mathbb{E}_{R}\left[ \log R(Z \mid A) \right] \nonumber\\
	& = \sum_{i=1}^N \sum_{q=1}^Q\mathbb{E}_{R}\left[ C_{iq} \log \mathcal{N}\left(Z_{i}; \mu_q, \sigma_{q}^2 I_p \right) \right] \\
	& \phantom{= } - \sum_{i=1}^N \mathbb{E}_{R}\left[\log \mathcal{N}\left(Z_{i}; \mu_{\phi_Z}(A)_{i}, \sigma_{\phi_Z}^2(A)_{i} I_p \right) \right]\nonumber\\
	&= - \sum_{i=1}^N \sum_{q=1}^Q \tau_{iq} \KL \left( \mathcal{N}\left( \mu_{\phi_Z}(A)_{i}, \sigma_{\phi_Z}^2(A)_{i} I_p \right) \mid \mid \mathcal{N}\left(\mu_q, \sigma_{q}^2 I_p \right) \right) \nonumber\\
	&= -  \sum_{i=1}^N \sum_{q=1}^Q \tau_{iq} \underbrace{\left[ \log \frac{\sigma_q^{p}}{\sigma_{\phi_Z}(A)_{i}^{p}} - \frac{p}{2} + \frac{ p \sigma_{\phi_Z}^2(A)_{i} + \| \mu_{\phi_Z}(A)_{i} - \mu_q \|_2^2 }{ 2 \sigma_q^{2}} \right]}_{\KL_{iq}^Z\left(\mu_{\phi_Z}(A)_{i}, \sigma_{\phi_Z}(A)_{i}, \mu_q, \sigma_{q}\right)} \nonumber\\
	&= -  \sum_{i=1}^N \sum_{q=1}^Q \tau_{iq} \KL_{iq}^Z\left(\mu_{\phi_Z}(A)_{i}, \sigma_{\phi_Z}(A)_{i}, \mu_q, \sigma_{q}\right) .
\end{align}
Symmetrically, the term regarding the edge positions is obtained as follow:
\begin{align*}
	&\mathbb{E}_{R}\left[ \log p(Y \mid A, C, m, s)\right] -\mathbb{E}_{R}\left[ \log R(Y \mid A, W) \right] \nonumber\\
	= & \sum_{i,j=1}^N \sum_{q,r=1}^Q\mathbb{E}_{R}\left[ A_{ij} C_{iq} C_{jr} \log \mathcal{N}\left(Y_{ij}; m_{qr}, s_{qr}^2 I_K \right) \right] \\
	& - \sum_{i,j=1}^N \mathbb{E}_{R}\left[A_{ij} \log \mathcal{N}\left(Y_{ij}; \mu_{\phi_Y}(W_{ij}), \diag\left(\sigma_{\phi_Y}^2(W_{ij} ) \right)  \right) \right]\nonumber\\
	= & - \sum_{i,j=1}^N \sum_{q,r=1}^Q A_{ij} \tau_{iq}\tau_{jr} \KL \left( \mathcal{N}\left( \mu_{\phi_Y}(W_{ij}), \diag \left(\sigma_{\phi_Y}^2(W_{ij}) \right) \right) \mid \mid \mathcal{N}\left( m_{qr}, s_{qr}^2 I_K \right) \right) \nonumber\\
	= & - \sum_{i,j=1}^N \sum_{q,r=1}^Q A_{ij} \tau_{iq}\tau_{jr} \KL_{ij qr}^Y\left( \mu_{\phi_Y}(W_{ij}),\sigma_{\phi_Y}(W_{ij}) , m_{qr}, s_{qr}  \right),
\end{align*}
where
\begin{equation*}
	\begin{split}
		\KL_{ij qr}^Y & \left( \mu_{\phi_Y}(W_{ij}), \sigma_{\phi_Y}(W_{ij}) , m_{qr}, s_{qr}  \right) \\
		& =  K \log s_{qr} - \sum_{k=1}^K \log \sigma_{\phi_Y}(W_{ij})_{k} - \frac{K}{2}  \\
		&\phantom{ = }+  \frac{\sum_{k=1}^K \sigma_{\phi_Y}^2(W_{ij})_{k} + \| \mu_{\phi_Y}(W_{ij}) -  m_{qr} \|_2^2}{ 2 s_{qr}^2}.
	\end{split}
\end{equation*}
\section{Inference}

\subsection{Optimisation}
In this section, we provide the optimisation steps to recover the parameters maximising the ELBO. To begin with, let us recall the ELBO expression:
\begin{align*}
	\mathscr{L}&(\tau, \phi_Z, \phi_Y; \pi, \mu, \sigma, \kappa, m, s, \alpha, \rho ):= \mathscr{L}(R(\cdot), \Theta) \\ 
	=&\sum_{i,j} \bigl\{ A_{ij}\mathbb{E}_{R}\left[\log p_{ij} \right] + (1 - A_{ij})\mathbb{E}_{R}\left[\log \bigl( 1- p_{ij} \bigr) \right]\bigr\} \\
	& + \sum_{i,j} A_{ij}\mathbb{E}_{R}\left[ \log \frac{ M_{ij}!}{\prod_{v=1}^V (W_{ij}^v)!} \prod_{v=1}^V (w_{ij}^v)^{W_{ij}^v} \right] \\
	&- \sum_{i=1}^N \sum_{q=1}^Q \tau_{iq} \KL_{iq}^Z\left(\mu_{\phi_Z}(A)_{i}, \sigma_{\phi_Z}^2(A)_{i} I_p, \mu_q, \sigma_{q}^2 I_p \right)\\
	& - \sum_{i,j=1}^N \sum_{q,r=1}^Q A_{ij} \tau_{iq}\tau_{jr} \KL_{ij qr}^Y\left( \mu_{\phi_Y}(W_{ij}), \diag\left(\sigma_{\phi_Y}^2(W_{ij})\right) , m_{qr}, s_{qr}^2 I_K  \right) \\
	&+ \sum_{i=1}^N \sum_{q=1}^Q \tau_{iq} \log \frac{\pi_{q}}{\tau_{iq}}.
\end{align*}
\paragraph{Update of $\tau$}
First, we optimise the ELBO with respect to $\tau_{iq}$. Since $\tau_i \in \Delta_{Q - 1}$, the term $c_i(1- \sum_{q=1}^Q \tau_{iq})$ is added to the ELBO, giving the Lagrangian of the function. Thus, the derivative of the Lagrangian with respect to $\tau_{iq}$ gives:

\begin{equation*}
	\begin{split}
		\frac{\partial }{\partial \tau_{iq}} \mathcal{L}(R(\cdot); \Theta) =& - \KL_{iq}^Z - \sum_{j=1}^N \sum_{r=1}^Q \left\{ A_{ij} \tau_{jr} \KL_{ij qr}^Y + A_{ji} \tau_{jr} \KL_{ji rq}^Y  \right\}\\
		& + \log \frac{\pi_{q}}{\tau_{iq}} - 1 - c_{i}.
	\end{split}
\end{equation*}

Setting this partial derivative to zero gives:
\begin{align*}
	\log \tau_{iq} = \underbrace{- \KL_{iq}^Z - \sum_{j=1}^N \sum_{r=1}^Q \left\{ A_{ij} \tau_{jr} \KL_{ij qr}^Y + A_{ji} \tau_{jr} \KL_{ji rq}^Y  \right\} + \log \pi_{q} - 1}_{T_{iq}} - c_{i}.
\end{align*}
Thus  $ \tau_{iq}= e^{T_{iq}} e^{-c_{i}}$. Moreover, since $\sum_{q=1}^Q \tau_{iq} = 1$, we have $e^{c_i} = \sum_{q=1}^Q e^{T_{iq}}$. Therefore, $\tau_{iq} = e^{T_{iq}} / ( \sum_{q=1}^Q e^{T_{iq}} )$. The complete form is:
\begin{align}
	\tau_{iq} & =  \frac{\pi_q e^{- \KL_{iq}^Z - \sum_{j \neq i } \sum_{r=1}^Q \left( A_{ij} \tau_{jr} K_{ij,qr}^Y + A_{ji} \tau_{jr} \KL_{ji rq}^Y  \right)} }{\sum_{l=1}^Q \pi_l e^{-\KL_{il}^Z - \sum_{j \neq i } \sum_{l'=1}^Q \left( A_{ij} \tau_{jl'} K_{ij, ll'}^Y  + A_{ji} \tau_{jl'} \KL_{ji l'l}^Y \right)}}.
\end{align}

\paragraph{Update of $\pi$ }
Since $\pi \in \Delta_{Q - 1}$, this constraint is added to the function to obtain the lagrangian. This corresponds to adding the term $c\left(1-\sum_{q=1}^Q \pi_q \right)$ to the ELBO. Thus, the partial derivative of the Lagrangian $\mathcal{L}$ with respect to $\pi_q$ of the lagrangian is:
\begin{align*}
	\frac{\partial }{\partial \mu_q} \mathcal{L}(R(\cdot); \Theta, c) = \sum_{i=1}^N \frac{\tau_{iq}}{\pi_q} - c.
\end{align*}
Setting this to zero gives:
\begin{align*}
	\frac{1}{\pi_q} \sum_{i=1}^N \tau_{iq} = c.
\end{align*}
Multiplying by $\pi_q$ and summing over $q$ gives that $c=N$. Therefore, after plugging it back into the previous expression, the following holds: 
\begin{align}
	\pi_q& = \frac{1}{N} \sum_{i=1}^N \tau_{iq}.
\end{align}

\paragraph{Updates of $\mu_q$ and $\sigma_q$}\label{proof:Z_params_update} Taking the partial derivative of the ELBO with respect to $\mu_q$ gives the following:
\begin{align*}
	\frac{\partial }{\partial \mu_q} \mathscr{L}(R(\cdot); \Theta) & = - \sum_{i=1}^N \frac{\tau_{iq}}{2 \sigma_q^2}\left(2 \mu_q - 2 \mu_{\phi_Z}(A)_{i} \right).
\end{align*}
Therefore, setting this quantity to zero gives the following update for $\mu_q$:
\begin{align}
	\mu_q = \left(\sum_{i=1}^N \tau_{iq}\right)^{-1} \sum_{i=1}^N \tau_{iq}  \mu_{\phi_Z}(A)_{i}.
\end{align}

The partial derivate of the ELBO with respect to $\sigma_q$ is:
\begin{align*}
	\frac{\partial }{\partial \sigma_q} \mathscr{L}(R(\cdot); \Theta) & =  -  \sum_{i=1}^N  \tau_{iq}  \left( \frac{p}{\sigma_q} -  \frac{p \sigma_{\phi_Z}^2(A)_{i} + \| \mu_{\phi_Z}(A)_{i} - \mu_q \|_2^2}{2} \frac{2 \sigma_q}{\sigma_q^4} \right).
\end{align*}
Thus, the first order condition on $\sigma_q$ gives the following:
\begin{align}
	\frac{p}{\sigma_q} \sum_{i=1}^N  \tau_{iq} & = \frac{1}{\sigma_q^3} \sum_{i=1}^N \tau_{iq} \left(p \sigma_{\phi_Z}^2(A)_{i} + \| \mu_{\phi_Z}(A)_{i} - \mu_q \|_2^2 \right)\nonumber \\
	\sigma_q^2 & =  \left(p\sum_{i=1}^N \tau_{iq}\right)^{-1} \sum_{i=1}^N \tau_{iq} \left( p \sigma_{\phi_Z}^2(A)_{i} + \| \mu_{\phi_Z}(A)_{i} - \mu_q \|_2^2 \right).
\end{align}

\paragraph{Updates of $m$ and $s$}\label{proof:Y_params_update}
As in the previous sections, we optimise the ELBO with respect to $m$ and $s$ with the first order conditions. The partial derivate of $\mathscr{L}$ with respect to $m_{qr}$ is:
\begin{align*}
	\frac{\partial }{\partial m_{qr}} \mathscr{L}(R(\cdot); \Theta) & = - \frac{1}{{2 s_{qr}^2}} \sum_{i,j=1}^N  A_{ij}\tau_{iq}\tau_{jr}\bigl(2 m_{qr} - 2 \mu_{\phi_Y}(W_{ij}) \bigr).
\end{align*}
Therefore, setting this expression to zero gives the following update for $m_{qr}$:
\begin{align}
	m_{qr} = \left(\sum_{i,j=1}^N A_{ij}\tau_{iq}\tau_{jr} \right)^{-1} \sum_{i,j=1}^N A_{ij}\tau_{iq}\tau_{jr}  \mu_{\phi_Y}(W_{ij}).
\end{align}

The partial derivate of the ELBO with respect to $s_{qr}$ is:
\begin{align*}
	\frac{\partial }{\partial s_{qr}} \mathscr{L}(R(\cdot); \Theta) & =  -  \sum_{i=1}^N   A_{ij}\tau_{iq}\tau_{jr}  \bigg( \frac{K}{s_{qr}} \\
	& \phantom{ =  -  \sum_{i=1}^N   A_{ij}\tau_{iq}\tau_{jr}  ( }-  \frac{\sum_{k=1}^K \sigma_{\phi_Y}^2(W_{ij})_{k}  + \| \mu_{\phi_Y}(W_{ij}) - m_{qr} \|_2^2}{ s_{qr}^3 } \bigg).
\end{align*}
Thus, the first order condition on $s_{qr}$ gives the following:
\begin{equation*}
	s_{qr}^2 K \sum_{i,j=1}^N A_{ij}\tau_{iq}\tau_{jr} = \sum_{i,j=1}^N A_{ij}\tau_{iq}\tau_{jr} \left[\sum_{k=1}^K \sigma_{\phi_Y}^2(W_{ij})_{k}  + \| \mu_{\phi_Y}(W_{ij}) - m_{qr} \|_2^2 \right].
\end{equation*}
Hence, the update of $s_{qr}^2$ is given by:
\begin{equation*}
	\begin{split}
		s_{qr}^2 = \left( K \sum_{i,j=1}^N A_{ij}\tau_{iq}\tau_{jr}\right)^{-1} \sum_{i,j=1}^N A_{ij}\tau_{iq}\tau_{jr} \bigg[&\sum_{k=1}^K \sigma_{\phi_Y}^2(W_{ij})_{k} \\
		&  + \| \mu_{\phi_Y}(W_{ij}) - m_{qr} \|_2^2 \bigg].
	\end{split}
\end{equation*}

\subsection{Derivation of the selection model criterion}
\begin{proof}[Proof of Proposition \ref{prop:IC2L}]\label{proof:IC2L}
	\begin{sloppypar}
		Assuming a fully factorised prior distribution $p(\kappa, \pi, \mu, \sigma, m, s, \rho, \alpha)$ $= p(\kappa)  p(\pi) p(\mu) p( \sigma) p(m) p(s) p(\rho) p(\alpha)$, Lemma 3.1 in \cite{biernacki2000assessing} can be directly extended to our case to decompose the integral in \eqref{eq:int_ic2l}:
	\end{sloppypar}\noindent
	\begin{align} \label{eq:ic2l_decomposed}
		\log p(A,W,Z,Y,C \mid \mathcalzapf{M}, Q, K, P )   & = \log p(A \mid Z,  \mathcalzapf{M}) + \log p (Z \mid C, \mathcalzapf{M}, Q, P) \nonumber \\
		& + \log p(W \mid A, Y,  \mathcalzapf{M}) + \log p(Y \mid A,  C, \mathcalzapf{M}, K ) \nonumber\\
		& + \log p(C \mid  \mathcalzapf{M}, Q).
	\end{align} 
	\begin{sloppypar}
		Unfortunately, this expression cannot be computed since each term requires an integral with respect to the corresponding parameter. For instance, $p(A \mid  Z,  \mathcalzapf{M} )$ cannot be integrated analytically because of the logistic link function. Fortunately, a BIC-like approximation can be derived for $p(A | Z,  \mathcalzapf{M})$, $p(Z | C, \mathcalzapf{M}, Q, P)$, $p(Y | A, C, \mathcalzapf{M}, K )$ and  $p(W | A, Y,  \mathcalzapf{M}) $. For instance, $p(A | Z,  \mathcalzapf{M})$ can be approximated by:
	\end{sloppypar}\noindent
	\begin{align*}
		\log p(A \mid Z,  \mathcalzapf{M}) & = \log \int_{\kappa} p(A \mid \kappa, Z,  \mathcalzapf{M}) p(\kappa) d\kappa\\
		& \approx  \max_{\kappa} \log p( A \mid Z, \kappa, \mathcalzapf{M}) - \frac{\nu_{A, \mathcalzapf{M}}}{2} \log(n_{A}),
	\end{align*}
	where $\nu_{A, \mathcalzapf{M}}= 1$ denotes the number of free components in $\kappa$ and $n_{A} = N(N-1)$ denotes the number of observations in $A$. This can be applied to all terms except $p(C~\mid~\mathcalzapf{M}, Q)$ since the posterior cluster memberships probabilities $\tau_i$ can be on the boundary of the parameter space. Fortunately, this term can be computed analytically. By assuming a Dirichlet prior $\mathcal{D}_{Q}(\delta_1,\dots,\delta_Q)$ on the topic proportions $\pi$:
	\begin{align*}
		p(C \mid \mathcalzapf{M}, Q) & = \int p(C \mid \pi, \mathcalzapf{M}, Q) p(\pi) d\pi \\
		& = \frac{ \Gamma\left(\sum_{q=1}^Q \delta_q\right) }{ \prod_{q=1}^{Q} \Gamma\left(\delta_q\right)}    
		\frac{ \prod_{q=1}^{Q} \Gamma(n_q + \delta_q)}{ \Gamma\left(\sum_{q=1}^Q n_q + \delta_q\right) },
	\end{align*}
	where $n_q := \sum_{i=1}^N C_{iq}$. In this paper, we consider the non informative Jeffreys prior distribution ($\delta_q =1/2$), as in \cite{biernacki2000assessing} and \cite{daudin2008mixture}. Moreover, since $C_i$ is not available, we replace it with its maximum-a-posteriori estimate $\hat{C}_i$ where $\hat{C}_{iq} = 1$ if $q = \argmax (\tau_{i1}, \dots, \tau_{iQ})$, and $0$ otherwise, which in turn gives $\hat{n}_q := \sum_{i=1}^N \hat{C}_{iq}$. Using Stirling formula to approximate the Gamma function for a large value of $N$, we obtain:
	\begin{align}
		p(\hat{C} \mid \mathcalzapf{M}, Q) \approx p( \hat{C} \mid \hat{\pi},\mathcalzapf{M}, Q) - \frac{Q-1}{2} \log(N).
	\end{align}
	To conclude, since $Z$ and $Y$ are not available, we replace the missing data with their maximum-a-posteriori estimates $\hat{Z}$ and $\hat{Y}$. Denoting $IC2L$ the quantity $\log p(A,W, \hat{Z},\hat{Y}, \hat{C} \mid \mathcalzapf{M}, Q, K, P ) $, we obtain:
	\begin{align*}
		IC2L &= \max_{\theta} \log p(A,W,\hat{Z},\hat{Y}, \hat{C} \mid \theta, \mathcalzapf{M}, Q, K, P ) - \Omega(\mathcalzapf{M}, Q, K, P)  \\
		& =  \max_{\kappa} \log p(A \mid \hat{Z}, \kappa, \mathcalzapf{M}) - \frac{ 1 }{2} \log(N (N- 1)) \nonumber\\
		& + \max_{\mu, \sigma} \log p (\hat{Z} \mid \hat{C}, \mu, \sigma, \mathcalzapf{M}, Q, P) - \frac{Q P + Q}{2} \log(N)\nonumber \\
		& + \max_{\rho, \alpha }\log p(W \mid A, \hat{Y}, \rho, \alpha, \mathcalzapf{M}) - \frac{V L + K  L}{2} \log(M) \nonumber\\
		& + \max_{m, s} \log p(\hat{Y} \mid A, \hat{C}, m, s, \mathcalzapf{M}, K ) - \frac{ Q^2  K + Q^2  }{2} \log(M) \nonumber \\ 
		& + \max_{\pi }\log p(\hat{C} \mid \pi, \mathcalzapf{M}, Q) - \frac{ Q-1}{2} \log( N ),
	\end{align*}
	where
	\begin{equation*}
		\begin{split}
			\Omega(\mathcalzapf{M}, Q, K, P) = &\frac{1}{2}\log(N(N~-~1))\\
			& +\frac{Q (P + 2) - 1}{2} \log(N) \\
			& +  \frac{L (V + K)+ Q^2  (K + 1) }{2} \log(M).
		\end{split}
	\end{equation*}
\end{proof}

\section{Numerical experiments}
This section provides the entire ELBO evolution as well as the evolutions of the node ARI and the edge ARI, corresponding to the example in Section \ref{sec:main_features_deep_lptm}.
\begin{figure}[h]
	\centering
	\includegraphics[width=0.6\linewidth]{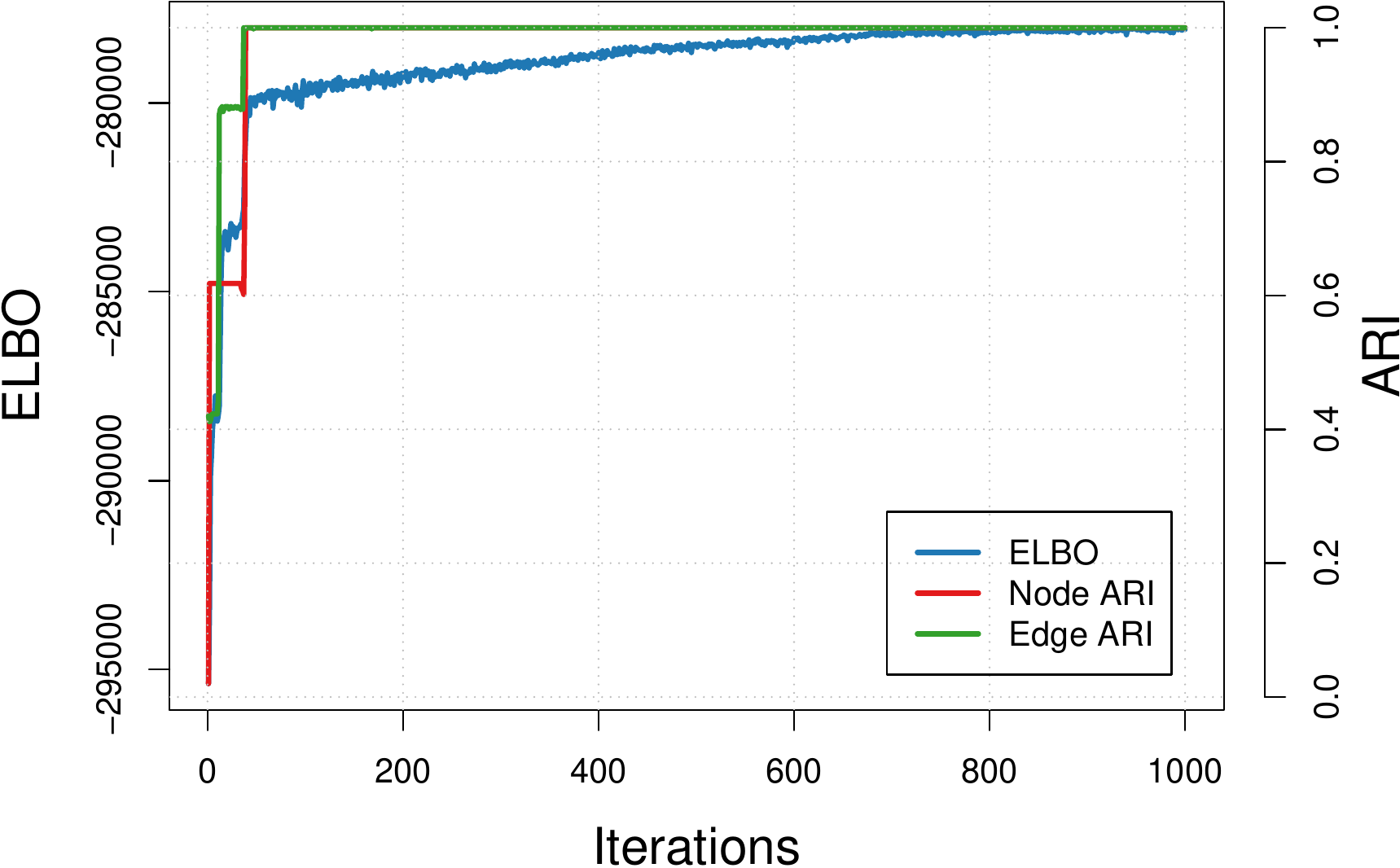}
	\caption{Evolution of the ELBO, as well as the nodes ARI during the optimisation. \hspace*{10cm}}
	\label{fig:example_sc_a_elbo_aris_1000}
\end{figure}

\section{Application to the analysis of the Enron email network}
\begin{figure}[ht]
	\centering
	\includegraphics[width=0.7 \textwidth]{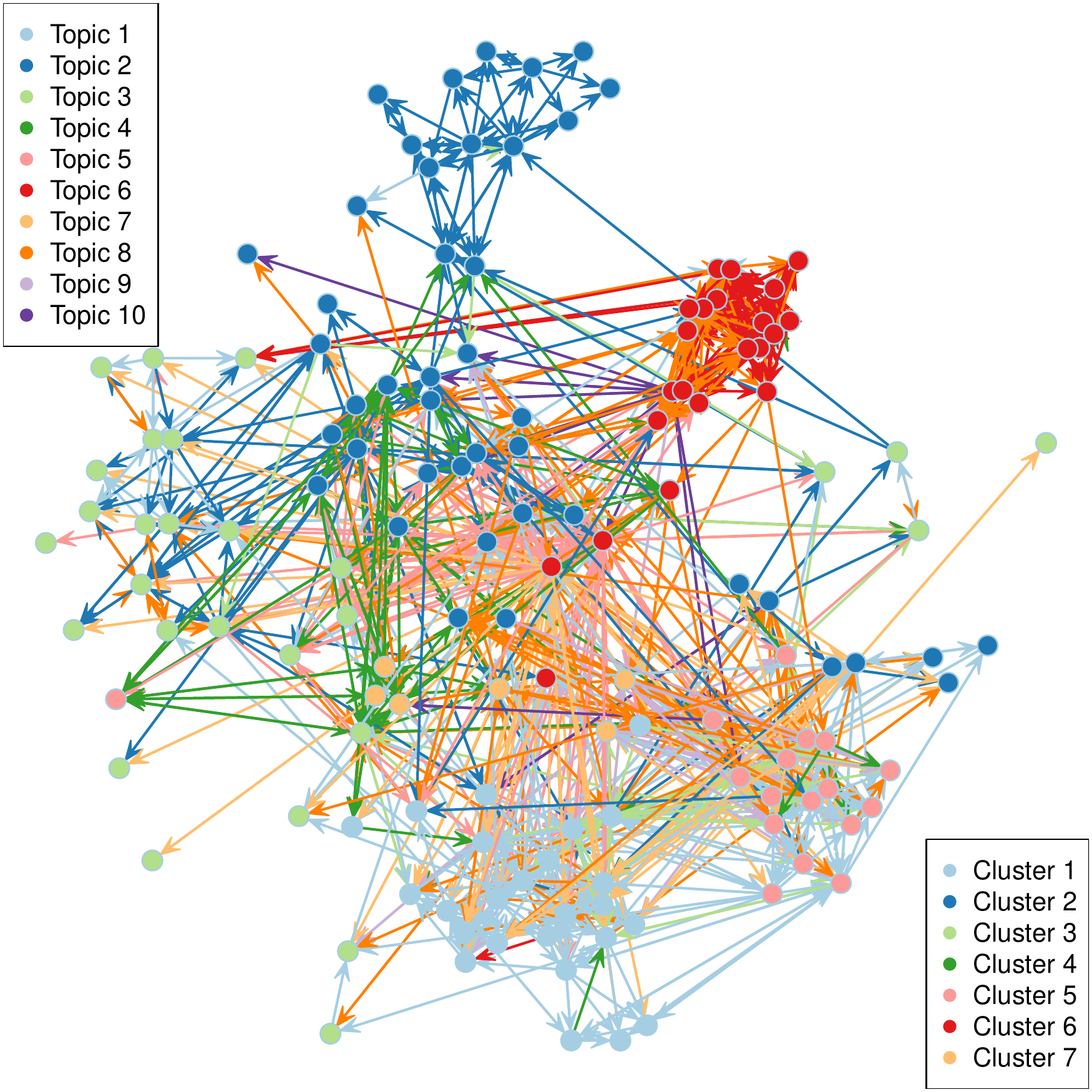}
	\caption{ETSBM clustering results on the Enron email network. The node positions were obtained using the Fruchterman-Reingold algorithm. \hspace*{10cm}}
	\label{fig:etsbm_enron}
\end{figure}

\begin{figure}[h]
	\centering
	\captionsetup[subfloat]{captionskip=0pt}
	\subfloat{\includegraphics[width=11cm]{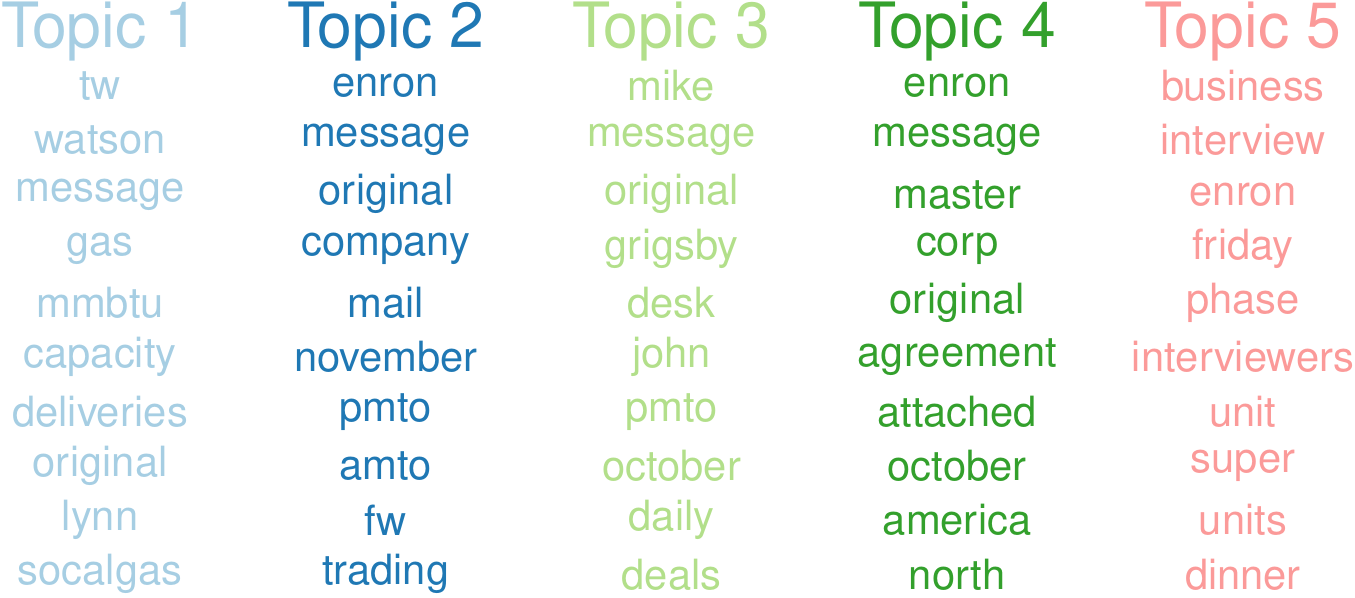}}\\
	\subfloat{\includegraphics[width=11cm]{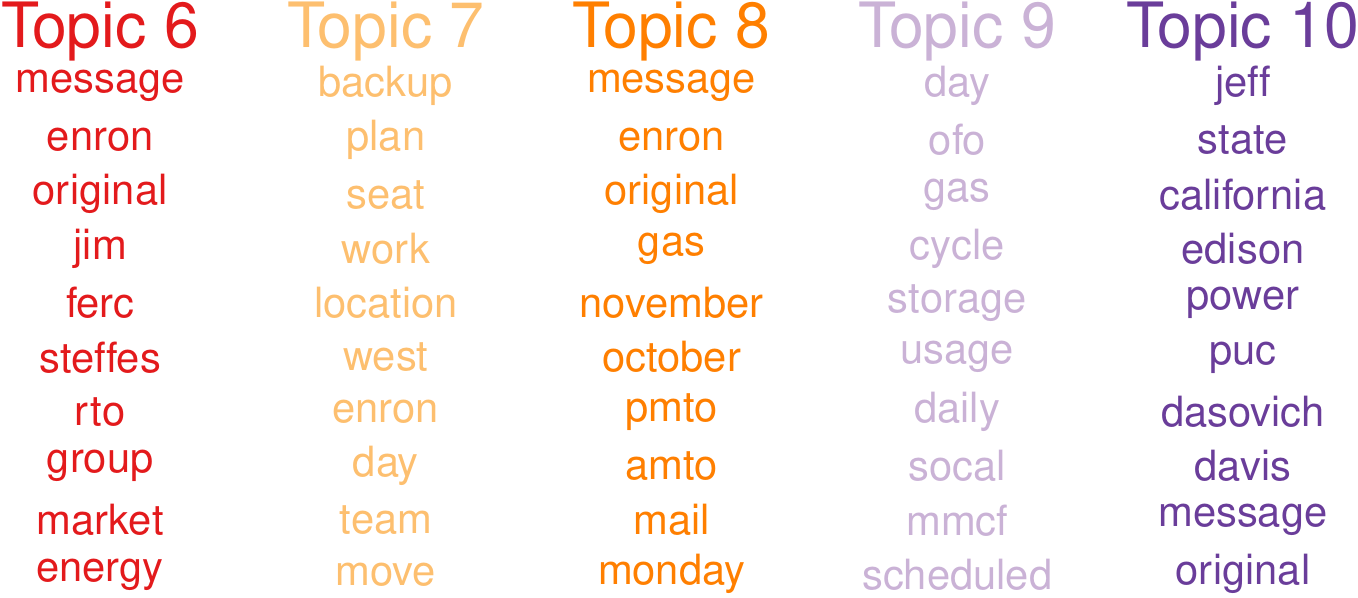}}
	\caption{{The 10 most probable words of each topic according to ETSBM.} \hspace*{10cm}}
	\label{fig:topics_etsbm_enron}
\end{figure}

To compare our results with a state-of-the-art approach, we provide an analysis of the Enron email network with ETSBM. The node positions displayed in Figure \ref{fig:etsbm_enron} were obtained using a  Freichterman-Reingold algorithm \cite{fruchterman1991graph} and the top words associated to ETSBM topics are provided in Figure \ref{fig:topics_etsbm_enron}.  

\newpage

\bibliography{document.bib}       

\begin{thebibliography}{10}
\providecommand \doibase [0]{http://dx.doi.org/}%

\bibitem{holland1983stochastic}
Holland PW, Laskey KB, Leinhardt S. Stochastic blockmodels: First steps. {\it
  Social networks.} 1983\string;5(2)\string:109--137.

\bibitem{snijders1997estimation}
Snijders TA, Nowicki K. Estimation and prediction for stochastic blockmodels
  for graphs with latent block structure. {\it Journal of classification.}
  1997\string;14(1)\string:75--100.

\bibitem{daudin2008mixture}
Daudin JJ, Picard F, Robin S. A mixture model for random graphs. {\it
  Statistics and computing.} 2008\string;18(2)\string:173--183.

\bibitem{hoff2002latent}
Hoff PD, Raftery AE, Handcock MS. Latent space approaches to social network
  analysis. {\it Journal of the american Statistical association.}
  2002\string;97(460)\string:1090--1098.

\bibitem{handcock2007model}
Handcock MS, Raftery AE, Tantrum JM. Model-based clustering for social
  networks. {\it Journal of the Royal Statistical Society: Series A (Statistics
  in Society).} 2007\string;170(2)\string:301--354.

\bibitem{snijders2011statistical}
Snijders TA. Statistical models for social networks. {\it Annual review of
  sociology.} 2011\string;37\string:131--153.

\bibitem{bouveyron2019model}
Bouveyron C, Celeux G, Murphy TB, Raftery AE. {\it Model-based clustering and
  classification for data science: with applications in R}. 50.
\newblock Cambridge University Press, 2019.

\bibitem{aljalbout2018clustering}
Aljalbout E, Golkov V, Siddiqui Y, Strobel M, Cremers D. Clustering with deep
  learning: Taxonomy and new methods. 2018.

\bibitem{kingma2014autoencoding}
Kingma DP, Welling M. Auto-Encoding Variational Bayes. 2014.

\bibitem{rezende2014stochastic}
Rezende DJ, Mohamed S, Wierstra D. Stochastic backpropagation and approximate
  inference in deep generative models. In: Proceedings of Machine Learning
  Research.  2014\string:1278--1286.

\bibitem{zhang2018network}
Zhang D, Yin J, Zhu X, Zhang C. Network representation learning: A survey. {\it
  IEEE transactions on Big Data.} 2018\string;6(1)\string:3--28.

\bibitem{kipf2016variational}
Kipf TN, Welling M. Variational graph auto-encoders. 2016.

\bibitem{kipf2016semi}
Kipf TN, Welling M. Semi-Supervised Classification with Graph Convolutional
  Networks. In:  2017.

\bibitem{henaff2015deep}
Henaff M, Bruna J, LeCun Y. Deep convolutional networks on graph-structured
  data. 2015.

\bibitem{pan2018adversarially}
Pan S, Hu R, Long G, Jiang J, Yao L, Zhang C. Adversarially Regularized Graph
  Autoencoder for Graph Embedding. In: IJCAI'18. AAAI Press
  2018\string:2609–2615.

\bibitem{mehta2019stochastic}
Mehta N, Duke LC, Rai P. Stochastic blockmodels meet graph neural networks. In:
  Proceedings of Machine Learning Research.  2019\string:4466--4474.

\bibitem{latouche2011overlapping}
Latouche P, Birmel{\'e} E, Ambroise C. Overlapping stochastic block models with
  application to the french political blogosphere. {\it The Annals of Applied
  Statistics.} 2011\string;5(1)\string:309--336.

\bibitem{liang2022deep}
Liang D, Corneli M, Bouveyron C, Latouche P. Deep latent position model for
  node clustering in graphs. In:  2022.

\bibitem{papadimitriou98latentsemantic}
Papadimitriou CH, Raghavan P, Tamaki H, Vempala S. Latent Semantic Indexing: A
  Probabilistic Analysis. {\it Journal of Computer and System Sciences.}
  2000\string;61(2)\string:217-235.

\bibitem{hofmann1999plsi}
Hofmann T. Probabilistic latent semantic analysis. In:  1999\string:289--296.

\bibitem{blei2003latent}
Blei DM, Ng AY, Jordan MI. Latent dirichlet allocation. {\it the Journal of
  machine Learning research.} 2003\string;3\string:993--1022.

\bibitem{blei2006correlated}
Blei DM, Lafferty J. Correlated topic models. {\it Advances in neural
  information processing systems.} 2006\string;18\string:147.

\bibitem{srivastava2017autoencoding}
Srivastava A, Sutton C. Autoencoding Variational Inference For Topic Models.
  In:  2017.

\bibitem{dieng2020topic}
Dieng AB, Ruiz FJR, Blei DM. Topic modeling in embedding spaces. {\it
  Transactions of the Association for Computational Linguistics.}
  2020\string;8\string:439--453.

\bibitem{mikolov2013efficient}
Mikolov T, Chen K, Corrado G, Dean J. Efficient estimation of word
  representations in vector space. 2013.

\bibitem{zhou2006probabilistic}
Zhou D, Manavoglu E, Li J, Giles CL, Zha H. Probabilistic models for
  discovering e-communities. In:  2006\string:173--182.

\bibitem{Rosenzvi2004author}
Rosen-Zvi M, Griffiths T, Steyvers M, Smyth P. The Author-Topic Model for
  Authors and Documents. In: UAI '04. AUAI Press 2004\string:487–494.

\bibitem{Pathak08socialtopic}
Pathak N, Delong C, Erickson K, Banerjee A. Social topic models for community
  extraction. 2008.

\bibitem{liu2009topic}
Liu Y, Niculescu-Mizil A, Gryc W. Topic-link LDA: joint models of topic and
  author community. In:  2009\string:665--672.

\bibitem{sachan2012using}
Sachan M, Contractor D, Faruquie TA, Subramaniam LV. Using content and
  interactions for discovering communities in social networks. In:
  2012\string:331--340.

\bibitem{bouveyron2018stochastic}
Bouveyron C, Latouche P, Zreik R. The stochastic topic block model for the
  clustering of vertices in networks with textual edges. {\it Statistics and
  Computing.} 2018\string;28(1)\string:11--31.

\bibitem{berge2019latent}
Berg{\'e} LR, Bouveyron C, Corneli M, Latouche P. The latent topic block model
  for the co-clustering of textual interaction data. {\it Computational
  Statistics \& Data Analysis.} 2019\string;137\string:247--270.

\bibitem{corneli2019dynamic}
Corneli M, Bouveyron C, Latouche P, Rossi F. The dynamic stochastic topic block
  model for dynamic networks with textual edges. {\it Statistics and
  Computing.} 2019\string;29(4)\string:677--695.

\bibitem{boutin2022embedded}
Boutin R, Bouveyron C, Latouche P. Embedded topics in the stochastic block
  model. {\it Statistics and Computing.} 2023\string;33(5)\string:1--20.

\bibitem{krivitsky2022statnet}
Pavel N.~Krivitsky MSH, Hunter DR, Butts CT, Klumb C, Goodreau SM, Morris M.
  Statnet: Tools for the Statistical Modeling of Network Data.

\bibitem{fruchterman1991graph}
Fruchterman TM, Reingold EM. Graph drawing by force-directed placement. {\it
  Software: Practice and experience.} 1991\string;21(11)\string:1129--1164.

\bibitem{kingma2019introduction}
Kingma DP, Welling M, others . An introduction to variational autoencoders.
  {\it Foundations and Trends{\textregistered} in Machine Learning.}
  2019\string;12(4)\string:307--392.

\bibitem{pytorch2019nips}
Paszke A, Gross S, Massa F, et al. PyTorch: An Imperative Style,
  High-Performance Deep Learning Library. In:  Wallach H, Larochelle H,
  Beygelzimer A, Alch\'{e}-Buc dF, Fox E, Garnett R. \kern-2pt, eds. {\it
  Advances in Neural Information Processing Systems 32}, , Curran Associates,
  Inc.,  2019\string:8024--8035.

\bibitem{kingma2014adam}
Kingma DP, Ba J. Adam: A method for stochastic optimization. 2014.

\bibitem{biernacki2000assessing}
Biernacki C, Celeux G, Govaert G. Assessing a mixture model for clustering with
  the integrated completed likelihood. {\it IEEE transactions on pattern
  analysis and machine intelligence.} 2000\string;22(7)\string:719--725.

\end{thebibliography}

\end{document}